\documentclass[10pt,twocolumn,letterpaper]{article}

\usepackage{iccv}              
\usepackage[accsupp]{axessibility}
\usepackage{overpic}
\usepackage{colortbl}
\usepackage{xcolor}  
\usepackage{tabularx}
\usepackage{arydshln}
\usepackage{pifont}
\usepackage{multirow}

\definecolor{ourscolor}{HTML}{D3D3D3}

\definecolor{LightBlue}{RGB}{30, 144, 255}

\def\methodname{DenseVLM}

\newcommand{\figref}[1]{Fig.~\ref{#1}}
\newcommand{\tabref}[1]{Tab.~\ref{#1}}
\newcommand{\eqnref}[1]{Eqn.~(\ref{#1})}
\newcommand{\secref}[1]{Sec.~\ref{#1}}

\newcommand{\cmark}{\ding{51}}%
\newcommand{\xmark}{\ding{55}}%

\newcommand{\myPara}[1]{\vspace{2pt}\noindent\textbf{#1}}

\newcommand{\tablestyle}[2]{\setlength{\tabcolsep}{#1}\renewcommand{\arraystretch}{#2}\centering\normalsize}

\definecolor{cvprblue}{rgb}{0.21,0.49,0.74}
\usepackage[pagebackref,breaklinks,colorlinks,allcolors=cvprblue]{hyperref}

\def\paperID{503} 
\def\confName{ICCV}
\def\confYear{2025}

\title{Unbiased Region-Language Alignment for Open-Vocabulary Dense Prediction}

\renewcommand{\thefootnote}{\fnsymbol{footnote}}

\author{Yunheng Li$^{1}$ ,
Yuxuan Li$^1$,
Quan-Sheng Zeng$^1$,
Wenhai Wang$^{3,4}$,
Qibin Hou$^{1,2,\dagger}$,
Ming-Ming Cheng$^{1,2}$ \\[0.5em]
$^1$VCIP, CS, Nankai University ~~~~~ $^2$NKIARI, Shenzhen Futian \\
$^3$OpenGVLab, Shanghai AI Laboratory ~~~~~ $^4$The Chinese University of Hong Kong \\
\textit{yunhengli@mail.nankai.edu.cn, yuxuan.li.17@ucl.ac.uk} \\
}

\begin{document}
\maketitle

%

\begin{abstract}
Pre-trained vision-language models (VLMs), such as CLIP, have demonstrated impressive zero-shot recognition capability, but still underperform in dense prediction tasks.
Self-distillation recently is emerging as a promising approach for fine-tuning VLMs to better adapt to local regions without requiring extensive annotations.
However, previous state-of-the-art approaches often suffer from significant `foreground bias', where models tend to wrongly identify background regions as foreground objects.
To alleviate this issue, we propose \methodname{}, a framework designed to learn unbiased region-language alignment from powerful pre-trained VLM representations. 
\methodname{} leverages the pre-trained VLM to retrieve categories for unlabeled regions and then decouples the interference between foreground and background features.
%
This separation ensures accurate region-category alignment while maintaining semantic distinctions during training.
We show that \methodname{} can directly replace the original VLM in open-vocabulary object detection and image segmentation methods, leading to notable performance improvements.
Furthermore, it exhibits promising zero-shot scalability when training on more extensive and diverse datasets.
Our code is publicly available \textcolor{blue}{\url{https://github.com/HVision-NKU/DenseVLM}}.
\end{abstract}

\footnotetext[2]{Corresponding Author. Email: houqb@nankai.edu.cn}

\section{Introduction}
Open-vocabulary dense prediction, primarily including object detection~\cite{Zareian_2021_CVPR,gu2021open,minderer2022simple,kuo2022f,jia2024mssd} 
and image segmentation~\cite{openseg,odise,ovseg,san,catseg,li2024cascade}, 
aims to identify regions or dense visual concepts of arbitrary categories as described by the text.
Benefiting from the powerful pre-trained Vision-Language Models (VLMs),
recent open-vocabulary approaches~\cite{maskclip1,Kim_2023_CVPR,san} for dense prediction have achieved great progress.
%

\begin{figure}[t]
    \centering
    \begin{overpic}[width=0.46\textwidth]{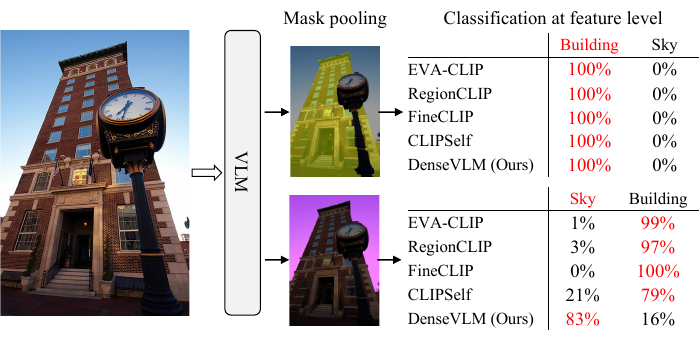}

    \end{overpic}
    \vspace{-12pt}
    \caption{{Illustration of foreground bias.}
    Previous methods~\cite{sun2023eva,zhong2022regionclip,wu2023clipself} often produce similar foreground predictions for background regions, 
    our approach effectively alleviates this issue.
    }
    \label{fig: thing_stuff}
    \vspace{-17pt}
  \end{figure}

Popular VLMs, such as CLIP~\cite{CLIP} and EVA-CLIP~\cite{sun2023eva}, have exhibited remarkable zero-shot recognition abilities for global image understanding~\cite{huang2024class,huang2025mindgappreservingcompensating,li2024advancing,li2024promptkd,fang2025aligning}. 
However, these models expose notable limitations in the understanding of local visual semantics, 
particularly in localizing and identifying small objects and background stuff~\cite{zhong2022regionclip,maskclip1}.
This limitation arises from the training manner of VLMs that align images with global text while neglecting the correspondences between local image regions and their specific text descriptions.
To alleviate this issue, some studies use region-text or pseudo region-text pairs~\cite{zhong2022regionclip,li2022grounded,zhang2022glipv2,liu2023grounding} but these methods are limited by the high annotation cost and lack scalability for open-world scenes.
In contrast, self-supervised approaches, such as CLIPSelf~\cite{wu2023clipself} and MaskEmbed~\cite{covert2024locality}, 
align region semantics using cropped image representations or reconstruct masked patch embeddings, respectively.
These self-distillation approaches, which do not rely on annotated data, 
offer flexibility and scalability across a variety of datasets.

\begin{figure*}[t]
    \centering
    \begin{overpic}[width=0.99\textwidth]{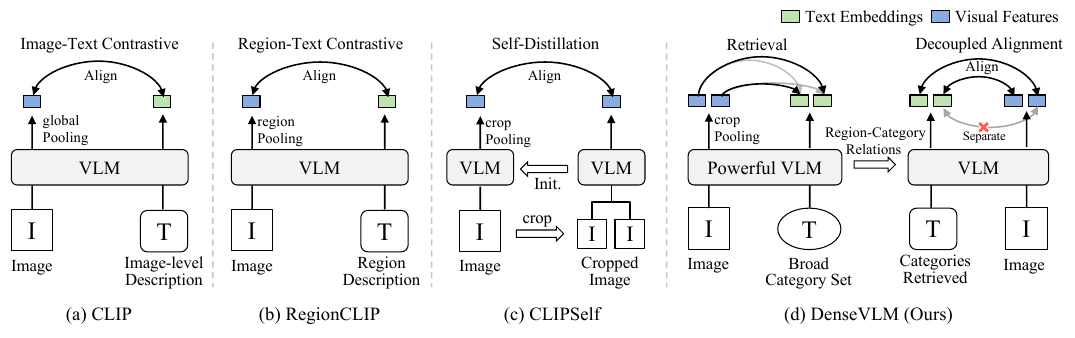}
    \end{overpic}
    \vspace{-10pt}
    \caption{{Comparison of different VLMs.}
    Unlike existing methods using (a) image-text contrastive learning~\cite{CLIP}, 
    (b) region-text contrastive learning~\cite{zhong2022regionclip} or (c) self-distillation~\cite{wu2023clipself}, 
    our method leverages powerful model representations for region-language alignment.
    }
    \label{fig: VLMs}
    \vspace{-12pt}
\end{figure*}

Despite the advances, previous VLMs~\cite{CLIP,sun2023eva,zhong2022regionclip}
pre-trained on image-text and region-text pairs tend to overemphasize the foreground objects at the expense of the background context.
This disproportionate focus results in a pronounced predilection for foreground object recognition 
and causes models to incorrectly associate background regions with foreground labels.
Consequently, in dense prediction tasks, these VLMs often misclassify background regions 
to co-occurring foreground classes---a phenomenon we term as `\emph{foreground bias}.'
To illustrate this issue,
we present an example comparing the classification results of popular VLMs~\cite{sun2023eva,zhong2022regionclip,wu2023clipself}, 
where features are extracted from regions using ground-truth masks.
As illustrated in \figref{fig: thing_stuff}, 
these models tend to confuse `sky' (a background class) with `building' (a foreground object).

%
%


To address this bias, we propose aligning foreground and background regions separately, ensuring explicit semantic separation through distinct category sets.
To achieve this, we introduce \methodname{}, an end-to-end framework designed for unbiased region-language alignment, as shown in \figref{fig: VLMs}(d).
Specifically, for unlabeled regions, we leverage a powerful pre-trained VLM that has learned robust semantic features from diverse data to retrieve relevant categories
without relying on paired data or self-distillation~\cite{sun2023eva, zhong2022regionclip, wu2023clipself}.
To ensure semantic diversity, \methodname{} incorporates a broad and comprehensive set of categories derived from large-scale datasets or predicted by generative models~\cite{Yue_2024_CVPR}.
A key feature of \methodname{} is its ability to classify regions as either foreground or background 
based on predefined sets for each category.
This classification way enables the decoupling of region features,
reducing interactions between foreground and background.
By decoupling these features, \methodname{} achieves independent region alignment of their respective categories while maintaining a semantic separation during training.
Furthermore, \methodname{} improves both efficiency and performance 
by directly extracting region features from the dense features of VLMs, 
avoiding traditional image cropping~\cite{zhu2024mrovseg,wu2023clipself}.

We evaluate the effectiveness of \methodname{} on several open-vocabulary benchmarks~\cite{wu2023clipself},
including object detection (box classification) and image segmentation (thing and stuff mask recognition).
\methodname{} is adaptable to various network architectures, such as ViTs~\cite{dosovitskiy2020image} and CNNs~\cite{lecun1989backpropagation}, 
and consistently outperforms other competing methods~\cite{CLIP,sun2023eva,wu2023clipself}.
%
Furthermore, \methodname{} has great scaling ability, 
showing promising performance improvement when scaling up the training set based on the SA-1B~\cite{kirillov2023segment} dataset.
For downstream tasks, \methodname{} improves the two-stage models~\cite{kuo2022f} of OV-COCO~\cite{chen2015microsoft} and OV-LVIS~\cite{Gupta_2019_CVPR} in open-vocabulary object detection
and achieves notable gains in open-vocabulary semantic segmentation over state-of-the-art methods such as SAN~\cite{san} and Cat-Seg~\cite{catseg}.
We summarize our contributions as follows.
\begin{itemize}[leftmargin=1em]
    \item We identify the foreground bias issue in existing VLMs and propose region-text alignment by incorporating explicit semantic structuring through category guidance.
    \item We propose \methodname{}, a region-language alignment framework that leverages a powerful VLM to retrieve categories for unlabeled regions and decouples foreground and background features to reduce foreground bias.
    \item Extensive experiments on dense prediction benchmarks show that our \methodname{} outperforms previous methods and exhibits promising scalability.
\end{itemize}

\section{Related work}

\myPara{Open-vocabulary dense prediction.}
Open-vocabulary dense prediction approaches aim to overcome the constraints of predefined categories, 
thereby enhancing their application in object detection~\cite{Zareian_2021_CVPR, gu2021open, minderer2022simple, kuo2022f} 
and image segmentation~\cite{openseg, odise, ovseg, san, li2024cascade,li2024clip}.
The success of pre-trained vision-language models like CLIP~\cite{CLIP}, has further inspired advancements in this area.
In open-vocabulary detection, recent studies~\cite{gu2021open,Wu_2023_CVPR} 
exploit the CLIP models to effectively identify novel objects.
Furthermore, several works~\cite{kuo2022f,CORA} construct object detectors that utilize frozen CLIP encoders, thereby reducing computational overhead while maintaining performance. 
For open-vocabulary segmentation, a common two-stage pipeline~\cite{maskclip, zegformer, odise, liu2023open} 
integrates a class-agnostic mask generator with a fixed CLIP encoder to achieve cross-modal alignment and mask classification.
Recent methods also investigate the use of shared frozen CLIP with side adapter networks~\cite{xu2023san} or the adoption of single-stage frameworks~\cite{fcclip}.
However, due to its training on image-text pairs, 
CLIP lacks precise local vision-language alignment, which is essential for dense prediction tasks.
Although recent studies
fine-tune CLIP for pixel-level~\cite{catseg,xie2023sed} or mask-level~\cite{jiao2024collaborative} alignment, 
they are constrained by the scarcity of densely labeled data.

\begin{figure*}[t]
    \centering
    \begin{overpic}[width=0.93\textwidth]{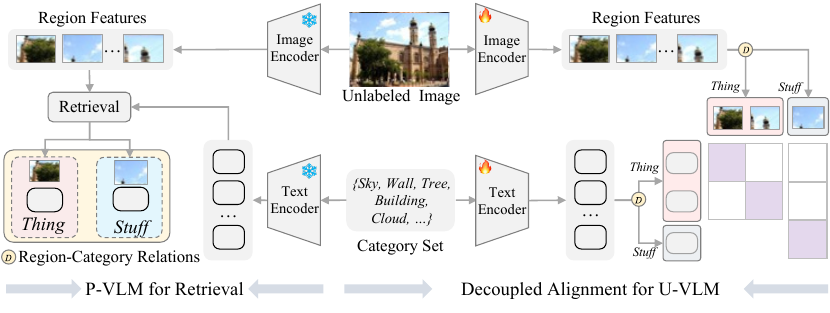}
    \put(2.5,28){$\mathcal{F}^1_{P}$}
    \put(7.5,28){$\mathcal{F}^k_{P}$}
    \put(14.5,28){$\mathcal{F}^{m \times n}_{P}$}

    \put(69.5,28){$\mathcal{F}^1_{U}$}
    \put(75,28){$\mathcal{F}^k_{U}$}
    \put(81.5,28){$\mathcal{F}^{m \times n}_{U}$}
  
    \put(26.1,17.5){$\mathcal{T}^1_{P}$}
    \put(26.1,13.7){$\mathcal{T}^2_{P}$}
    \put(26.1,8.5){$\mathcal{T}^D_{P}$}

    \put(2,20.5){$\mathcal{F}^u_{P}$}
    \put(4,13.0){$\mathcal{T}^u_{P}$}

    \put(16.5,20.5){$\mathcal{F}^v_{P}$}
    \put(14.5,13.0){$\mathcal{T}^v_{P}$}



    \put(69.8,17.7){$\mathcal{T}^1_{U}$}
    \put(69.8,13.9){$\mathcal{T}^2_{U}$}
    \put(69.8,8.5){$\mathcal{T}^D_{U}$}

    \put(82,22.8){$\mathcal{F}^u_{U}$}
    \put(100,22.8){$\mathcal{F}^v_{U}$}

    \put(81,17.5){$\mathcal{T}^1_{U}$}
    \put(81,13){$\mathcal{T}^u_{U}$}

    \put(81,8){$\mathcal{T}^v_{U}$}

    \end{overpic}
    \vspace{-10pt}
    \caption{{Overview of \methodname{}.}
    %
    %
    \methodname{} leverages the powerful VLM to \textbf{retrieve} categories for unlabeled regions 
    and distinguish between foreground and background.
    During VLM training, \methodname{} \textbf{decouples} interference between features of foreground and background regions, 
    \textbf{aligning} each region with its corresponding text embeddings.
    }
    \vspace{-12pt}
    \label{fig: framework}
  \end{figure*}

\myPara{Vision-language alignment at image and region levels.}
Pre-training visual-language models enable alignment between image and text representations~\cite{align,Vilbert,CLIP,jose2025dinov2,ma2024clip,chen2023vlp,wang2023large}. 
By using image-text pairs~\cite{schuhmann2022laion} as training data, methods such as CLIP~\cite{CLIP} and ALIGN~\cite{align} exhibit impressive zero-shot classification capabilities.
To improve the alignment of vision-language of dense features, 
some training-free studies~\cite{wang2025sclip,lan2024clearclip,lan2024proxyclip,wysoczanska2024clip} attempt to modify the output layers of CLIP. 
%
%
Recent methods~\cite{mukhoti2023open,ranasinghe2023perceptual, naeem2024silc} align visual patches with image-level text or 
learn local-global correspondence via self-distillation, both requiring extensive text-image pairs.

For precise local region alignment, researchers utilize annotations from visual grounding datasets~\cite{liu2023grounding} to train models on region-text alignment.
For example, some methods such as GLIP~\cite{li2022grounded,zhang2022glipv2} and Grounding DINO~\cite{liu2023grounding} learn region-language grounding from region-text pairs 
or generates pseudo region-text pairs like RegionCLIP~\cite{zhong2022regionclip}.
In the context of open-vocabulary detection and segmentation,
some works achieve dense visual and text alignment by using mask attention~\cite{maskclip,maskclip1,Kim_2023_CVPR,san} 
or fine-tuning CLIP~\cite{xie2023sed,catseg,jiao2023learning}.
%
However, these methods are constrained by the high cost of annotation, making large-scale deployment challenging.
To address the issue of annotated data scarcity, recent methods like CLIPSelf~\cite{wu2023clipself} use cropped images for semantic alignment,
while MaskEmbed~\cite{covert2024locality} leverages masked patch embeddings for feature reconstruction.
Despite these advances, the efficacy of self-distillation methods is constrained by the suboptimal performance of the teacher model
and is further compromised by foreground bias.
To overcome these limitations, we leverage a powerful VLM to retrieve categories for unlabeled regions 
while decoupling foreground and background features through textual category guidance.


%

\section{Method}

Our aim is to develop a region-language alignment model 
that can effectively align local visual and semantic features,
thereby enhancing the performance of VLMs in open-vocabulary dense prediction tasks.
To achieve this, it is crucial to alleviate the foreground bias problem prevalent in previous VLMs~\cite{CLIP,sun2023eva} 
that arises from training on image-text pairs.
Moreover, our approach seeks to receive better performance beyond the constraints of self-distillation~\cite{wu2023clipself}.

\subsection{VLM's representation}
\label{sec: vlm_representation}
VLMs are typically designed to learn both global visual 
and textual representations within a shared semantic space.
In ViT-based VLMs~\cite{openclip,sun2023eva}, dense visual features are extracted through residual attention blocks. 
Following~\cite{maskclip1,wu2023clipself}, we derive the dense image feature map $\mathcal F$ 
by modifying the final residual attention block,
\ie, removing the SoftMax operation and incorporating mapping layers.
For a set of categories $\{c^1, \dots, c^D\}$, 
where $D$ is the total number of categories, 
textual descriptions are generated using a prompt-template strategy~\cite{gu2021open}, 
like ``\textit{This is a photo of the ${c}$ in the scene}.''
These prompts are then encoded into text embeddings
$\mathcal T = \{\mathcal T^1, \ldots, \mathcal T^D\}$ by the text encoder.

\subsection{\methodname{} framework}
\label{sec: framework}

We propose \methodname{}, an end-to-end region-language alignment framework designed 
to mitigate foreground bias.
\methodname{} achieves this by precisely aligning unlabeled regions with their corresponding categories. 
As shown in~\figref{fig: framework}, the framework consists of two key components. 
First, it retrieves category semantics for region features using the P-VLMs (Powerful VLMs) with frozen weights.
Second, it decouples these region-language alignments into foreground and background to train U-VLMs (Unbiased VLMs) without foreground bias.
In particular, \methodname{} operates without requiring any annotations, 
relying on diverse category semantics from large-scale datasets~\cite{caesar2018coco,ade20k} 
or generating category sets from images using the NXTP~\cite{Yue_2024_CVPR}.

\myPara{Image patches to semantic regions.}
To achieve region alignment, the patch-level visual features from~\secref{sec: vlm_representation}
need to be transformed into semantic region features.
We adopt a strategy similar to~\cite{wu2023clipself}, dividing the dense feature map 
into an $m \times n$ grid of patches.
Unlike this approach, we refrain from directly cropping the input image, 
thereby enhancing both computational efficiency and representation effectiveness. 
In each training iteration, 
$m$ and $n$ are randomly selected from the set $\{2, \cdots, M\}$, 
where $M$ defaults to 6, allowing for varying patch grid sizes.
The semantic region representation $\{\mathcal F^1, \cdots, \mathcal F^{m \times n}\}$ 
is then extracted from dense feature map $\mathcal F$ via pooling (RoIAlign)~\cite{he2017mask}.
This patch sampling strategy effectively captures region features 
of both foreground objects and background scenes.
However, due to the foreground bias in VLMs~\cite{CLIP,sun2023eva,wu2023clipself}, we observe that background regions are usually misclassified as foreground classes, 
despite containing only a small proportion of foreground-related patches.

\myPara{Powerful VLM for retrieval regions.}
Building on the dense representation extracted from VLM described in~\secref{sec: vlm_representation} 
and its ability to map image patches to semantic regions,
the powerful VLM enables the extraction of both 
region features $\mathcal F_{P}=\{\mathcal F^1_{P}, \cdots, \mathcal F^{m \times n}_{P}\}$ 
and text embeddings $\mathcal T_{P}=\{\mathcal T^1_{P}, \ldots, \mathcal T^D_{P}\}$.
%
%
%
%
Next, unlabeled regions are retrieved and matched with the most relevant categories by computing the cosine similarity 
between the region features $\mathcal F_{P}$ and text embeddings $\mathcal T_{P}$. 
For a specific region $k$,
the cosine similarity between its features $\mathcal F^k_{P} = \mathcal F_{P}[k,:]$ and the text embeddings of all categories is computed as:
\begin{equation} \label{eq: cos}
    \cos(\mathcal F^k_{P}, \mathcal T^i_{P})=  
    \frac{\mathcal F^k_{P} \cdot \mathcal T^i_{P}}{\lVert \mathcal F^k_{P} \rVert \lVert \mathcal T^i_{P} \rVert}, \quad \forall i = 1, 2, \ldots, D
\end{equation}
where $\cdot$ denotes the dot product
and $\lVert \cdot \rVert$ represents the Euclidean norm.
The probability of associating this region with the categories is determined as follows:
\begin{equation} \label{eq: prob}
    p^k(y=c|\mathcal F^k_{P},\mathcal T_{P}) =  \frac{\exp( \cos(\mathcal F^k_{P},  \mathcal T_{P}^c)/\tau) }{\sum_{j=1}^D \exp( \cos(\mathcal F^k_{P}, \mathcal T_{P}^j)/\tau)},
\end{equation}
%
where $\tau = 0.01$ is a temperature hyperparameter.

The use of random grids to extract region features often introduces uncertainty 
in fully covering a single object, 
especially in scenarios with multiple objects.
This uncertainty significantly affects the precision of the region-category alignment. 
To mitigate this issue, we leverage a region denoising scheme by discarding any region whose matching probability falls below a threshold $\theta$, which is set to 0.3 by default.
Therefore, this retrieval and denoising process yields a more precise and reliable alignment of regions with their corresponding categories.
The optimal category for each region $k$, denoted as \(c^k\), is determined by:
$c^k = \operatorname{argmax}(p^k)$,
where each region-category relation is represented as $(k, c^k)$.
This alignment is fundamental for the subsequent decoupled alignment process.

\begin{table*}[t]
  \centering
  \tablestyle{7.5pt}{1.0}
  \centering
  \small
  \begin{tabular}{l  cc cc cc cc cc cc}
  \toprule
  & \multicolumn{6}{c}{COCO}& \multicolumn{6}{c}{ADE20K}\\
  \cmidrule(lr){2-7}
  \cmidrule(lr){8-13}
  & \multicolumn{2}{c}{\textbf{Boxes}}  &\multicolumn{2}{c}{\textbf{Masks-T}}&\multicolumn{2}{c}{\textbf{Masks-S}} & \multicolumn{2}{c}{\textbf{Boxes}}  &\multicolumn{2}{c}{\textbf{Masks-T}}&\multicolumn{2}{c}{\textbf{Masks-S}} \\
  \multirow{1}{*}{Method}&Top1&Top5&Top1&Top5& Top1&Top5&Top1&Top5&Top1&Top5& Top1&Top5 \\
  \midrule
  OpenCLIP~\cite{openclip}& 49.8& 74.3& 51.9& 72.2& 29.2& 54.9& 28.4& 54.1& 29.6& 53.4& 37.9& 66.6 \\
  DFN~\cite{fang2023data}& 38.3& 65.0& 31.0& 57.0& 26.4& 54.9& 30.6& 57.9& 24.2& 49.9& 32.2& 57.7 \\
  SigLIP~\cite{siglip}& 39.9& 61.4& 40.4& 60.1& 30.3& 56.4& 25.9& 49.2& 27.3& 47.6& 34.5& 57.3 \\
  EVA-CLIP~\cite{sun2023eva}& 44.3& 68.7& 44.7& 66.0& 26.2& 51.9& 33.0& 57.6& 33.9 & 56.2& 36.2& 62.3\\
  RegionCLIP$^\dagger$~\cite{zhong2022regionclip}& 68.5& 89.5& 60.7& 84.3& 22.0& 53.5& 43.2& 72.2& 34.0& 62.6& 37.7& 68.6\\
  FineCLIP$^\dagger$~\cite{jing2025fineclip}& 64.7& 86.1& 62.5& 80.9& 36.9& 70.3& 43.9& 71.2 & 45.5& 68.6& 46.0& 74.8\\
  CLIPSelf$^\dagger$~\cite{wu2023clipself}& 69.1& 88.2& 66.7& 83.0& 41.7& 75.2& 48.1& 77.7& 47.5& 74.2& 53.7& 82.8\\
  \rowcolor[HTML]{EFEFEF}\methodname{}$^\dagger$ (Ours)& \textbf{72.3}& \textbf{89.9}& \textbf{70.1}& \textbf{84.4}& \textbf{44.9}& \textbf{76.4}& \textbf{51.0}& \textbf{81.8}& \textbf{49.3}& \textbf{76.5}& \textbf{57.0}& \textbf{84.0} \\ 
  \bottomrule
  \end{tabular}
  \vspace{-5pt}
\caption{
{Comparisons of different models on dense representation}.
We report the Top1 and Top5 mean accuracy on classifying boxes and panoptic masks (thing and stuff).
$^\dagger$ denotes models trained on the COCO and evaluated in a zero-shot setting on the ADE20K dataset.
}
\label{tab: compare_others}
\end{table*}

\begin{figure*}[t]
    \centering
    \vspace{-3pt}
    \begin{overpic}[width=0.98\textwidth]{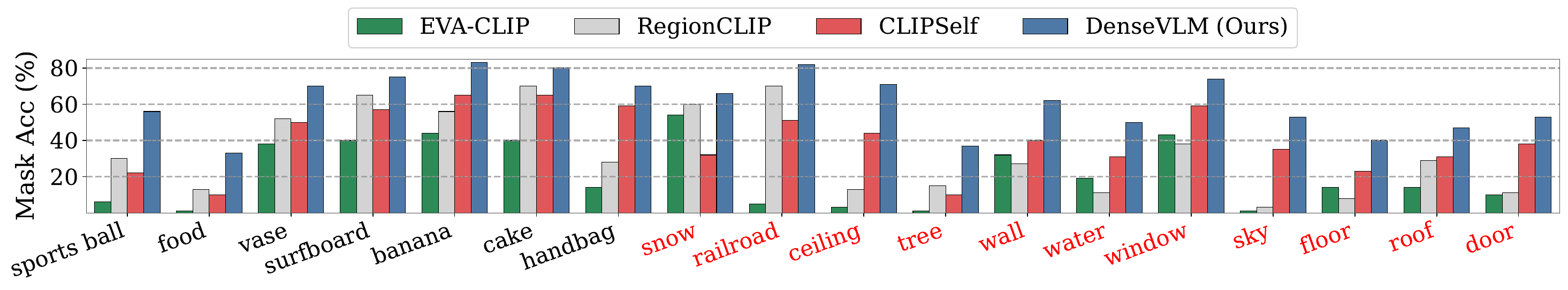}
  
    \end{overpic}
    \vspace{-8pt}
    \caption{
    Mask accuracy comparison across categories in COCO dataset.
    Our method achieves notable improvements, especially in addressing foreground bias.
    The foreground categories are shown in black, and the background categories are highlighted in \textcolor{red}{red}.
    }
    \label{fig: acc_comparison}
    \vspace{-15pt}
  \end{figure*}

\myPara{Decoupled region alignment to train VLMs without foreground bias.}
Upon the established region-category relations, we perform alignment between the region and text representations to train the U-VLM.
A straightforward approach would be to directly align region features and text embeddings of their corresponding categories
while maximizing the separation of unrelated categories. 
However, since the U-VLM directly inherits from the P-VLM, our experimental results in~\secref{sec: ablation_study} reveal that this method exacerbates foreground bias,
leading to improved foreground detection but limited gains in background recognition.

To mitigate this issue, we propose a decoupled alignment strategy that separates the alignment process for foreground and background regions.
Specifically, we denote the region features of the training U-VLM as $\mathcal F_{U}$, using the same partitioning grid as the P-VLM.
The text embeddings are represented as $\mathcal T_{U}$.
By leveraging the region-category relations $(k, c^k)$ retrieved by the P-VLM, 
we establish a one-to-one mapping for the U-VLM to associate the region features and their corresponding category embeddings.
To distinguish the semantic regions of foreground and background, we decouple these region-category relations into two distinct groups, 
following the predefined category sets: foreground objects 
\textit{Thing} ($\mathcal U$) and background regions \textit{Stuff} ($\mathcal V$).
%
By selectively contrasting against text embeddings related to irrelevant categories, 
we guide the model to focus more on the relevant background regions, 
reducing the impact of irrelevant foreground categories. 
This selective contrast helps the model capture the distinctive characteristics of background regions, 
leading to a more accurate separation between foreground and background.
The alignment process can be effectively achieved by maximizing the cosine similarity for the region features and the text embeddings.
According to~\eqnref{eq: prob}, 
when $c^k \in \mathcal{V}$, the probability $q^k$ for a specific region is calculated as: 
\begin{equation} \label{eq: q_s}
    q^k =  \frac{\exp( \cos(\mathcal F^k_{U},  \mathcal T_{U}^c)/\tau) }
    {\sum_{j=1}^{\mathcal{U}\cup \mathcal{V}} \exp( \cos(\mathcal F^k_{U}, \mathcal T_{U}^j)/\tau)}.
\end{equation}
Similarly, when $c^k \in \mathcal{U}$, the probability $\tilde q^k$ is computed as: 
\begin{equation} \label{eq: q_t}
    \tilde q^k = \begin{cases} \frac{\exp( \cos(\mathcal F^k_{U},  \mathcal T_{U}^c)/\tau) }
    {\sum_{j=1}^{\mathcal{U}} \exp( \cos(\mathcal F^k_{U}, \mathcal T_{U}^j)/\tau)} &  \text{if} ~ c \in \mathcal{U},
    \\
0 & \text{otherwise}.
\end{cases}
\end{equation}
%
%

\myPara{End-to-end optimization.}
As in~\eqnref{eq: q_s} and~\eqnref{eq: q_t}, we also compute $p^k$ and $\tilde p^k$ for P-VLM. 
The proposed method \methodname{} supports end-to-end training through KL-divergence optimization
as follows.
\begin{align}
    \label{eq:loss}
    \mathcal{L}^k=
    \begin{cases}
        \operatorname{KL}(p^k || q^k)  & \text{if}~ c^k \in \mathcal{U}, \\
        \operatorname{KL}(\tilde{p}^k || \tilde{q}^k)  & \text{otherwise}.
        \end{cases}
\end{align}
The overall loss for each image is computed as $\mathcal{L} = \frac{1}{m' \times n'}\sum^{m' \times n'}_{k} \mathcal{L}^k$, 
where the sum is taken over all regions, excluding those removed by the region denoising scheme.

\section{Experiments}
\label{sec: experiments}

\begin{table*}[!t]
    \tablestyle{5.5pt}{1.0}
    \centering
    \small
    \begin{tabular}{c c c c c cc cc cc}
    \toprule
    \multirow{1}{*}{VLMs}  &\multirow{1}{*}{Region} &\multirow{1}{*}{Alignment}&\multirow{1}{*}{GPU Memory} &\multirow{1}{*}{Time Overhead}  &\multicolumn{2}{c}{\textbf{Boxes}}  &\multicolumn{2}{c}{\textbf{Masks-T}}&\multicolumn{2}{c}{\textbf{Masks-S}} \\
    Frozen \& Training&Cropping&Strategy&(per card)&(per epoch)&Top1&Top5&Top1&Top5& Top1&Top5\\
    \midrule
    ViT-B/16 \& ViT-B/16 & Images& Features KD& 37G& 25min& 69.1& 88.2& 66.7& 83.4& 41.7& 75.2 \\
    ViT-L/14 \& ViT-B/16  & Images& Features KD& 39G& 37min& 24.2& 52.4& 23.4& 51.1& 10.1& 39.1 \\
    ViT-L/14 \& ViT-B/16  & Images & Logics KD & 39G& 55min& 72.2& 89.8& 68.8& 83.8& 42.6& 75.2 \\
    \rowcolor[HTML]{EFEFEF} ViT-L/14 \& ViT-B/16  & Features & \methodname{} & 39G& \textbf{23min}& \textbf{73.4}& \textbf{90.5}& \textbf{71.0}& \textbf{84.8}& \textbf{45.6}& \textbf{77.8} \\
    \bottomrule
    \end{tabular}
  \vspace{-5pt}
  \caption{
  Results and comparisons of various frameworks, including target VLMs for aligning representations, 
  region cropping of frozen VLMs, and optimization strategies for region-language alignment,
  with GPU memory efficiency and time overhead.
  All models are trained on four A40 GPUs,
  with each epoch containing 118$k$ images.
  }
  \label{tab: strategy}
\vspace{-13pt}
\end{table*}

\subsection{Benchmarks}
\label{sec: benchmarks}

\myPara{Experiment settings.}
To verify the effectiveness of the proposed \methodname{}, we perform experiments on 
dense prediction tasks using the COCO panoptic~\cite{lin2014microsoft} val2017 split and ADE20K panoptic~\cite{zhou2017scene} val split. 
Following CLIPSelf~\cite{wu2023clipself}, we evaluate box classification using pooled box features (\textbf{Boxes}) 
and mask classification with pooled mask features, 
distinguishing foreground objects (\textbf{Masks-T}) from background content (\textbf{Masks-S}). 
This process follows a similar procedure as illustrated in~\figref{fig: thing_stuff},
where we use ground truth annotations to extract local features 
and evaluate the local classification.
Results are reported in Top-1 and Top-5 mean accuracy across all experiments.

\myPara{Implementation details.}
We employ the ViT-L/14 model from CLIPSelf~\cite{wu2023clipself} as the powerful P-VLM 
and ViT-B/16 from EVA-CLIP~\cite{sun2023eva} as the training U-VLM.
To enhance computational efficiency, the powerful VLM are kept frozen, 
and only the image encoder of the  U-VLM is trained with pre-extracted text embeddings.
Considering the practical application of the downstream tasks and the trade-off between performance and efficiency,
we resize the input images to a uniform resolution of $512 \times 512$ pixels.
The models are trained for 6 epochs using the AdamW~\cite{adamw} optimizer with a weight decay of 0.1.
The initial learning rate is set to 1$\times10^{-5}$ with a cosine decay~\cite{loshchilov2017sgdr} schedule.

\subsection{Comparison with other VLMs}
\label{exp: other_vlms}

\myPara{Quantitative evaluation.}
We perform a comprehensive quantitative evaluation of dense representations 
across multiple VLMs on the COCO Panoptic~\cite{lin2014microsoft} and ADE20K Panoptic~\cite{zhou2017scene} datasets. 
%
%
As shown in \tabref{tab: compare_others}, while previous methods~\cite{openclip,fang2023data,siglip,sun2023eva} achieve strong zero-shot image classification performance, 
their performance in region recognition is notably suboptimal. 
For instance, EVA-CLIP~\cite{sun2023eva} achieves only a Top-1 box classification accuracy of 44.3\% on COCO 
and 33.0\% on ADE20K.
Although RegionCLIP~\cite{zhong2022regionclip}, which is trained on region-text pairs, shows improved performance on COCO, 
but fails to generalize to datasets such as ADE20K.
Moreover, FineCLIP~\cite{jing2025fineclip} and CLIPSelf~\cite{wu2023clipself}, which incorporate self-distillation, 
achieve higher Mask-T classification accuracy but perform poorly in Mask-S classification. 
In contrast, our method, \methodname{}, notably outperforms these models,
achieving a 4.3\% improvement in Top-1 Mask-T accuracy and a 3.9\% improvement in Top-1 Mask-S accuracy on COCO,
highlighting its effectiveness in dense prediction tasks.

\figref{fig: acc_comparison} shows a comparison of mask accuracy across different categories, 
further demonstrating the effectiveness of \methodname{} in region-text alignment.
Notably, our method notably improves accuracy in background categories such as `sky' and `wall', 
which can mitigate foreground bias.

  \begin{figure}[t]
    \centering
    \vspace{-5pt}
    \begin{overpic}[width=0.48\textwidth]{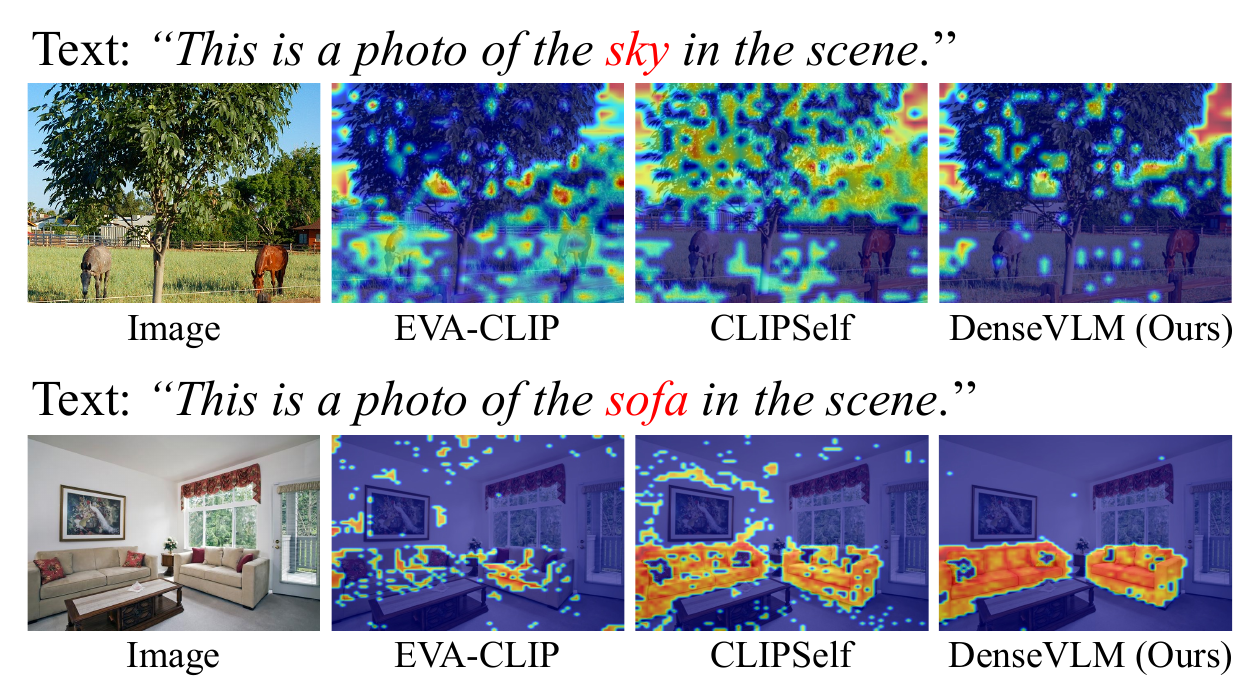}
  
    \end{overpic}
    \vspace{-20pt}
    \caption{
    Comparing the alignment effect of our \methodname{} with other methods through
    visualizations of cosine similarity maps between visual features and text embeddings.
    }
    \label{fig: cos_features}
    \vspace{-15pt}
  \end{figure}

\myPara{Qualitative results.}
%
%
We visualize attention maps using cosine similarity for text-described object categories.
As shown in \figref{fig: cos_features}, \methodname{} achieves more accurate and complete object localization than EVA-CLIP and CLIPSelf.
Moreover, it better preserves semantic separation, reducing interference from other objects.

\subsection{Ablation study}
\label{sec: ablation_study}

In our ablation study, \methodname{} is trained on the unlabeled images from the COCO train2017 split 
using 4$\times$A40 GPUs and evaluated on the val2017 split.
The experiments in~\secref{sec: ablation_study} utilize the ViT-B/16 model from EVA-CLIP~\cite{sun2023eva} 
for its superior efficiency and capacity.

\myPara{Framework design exploration.}
As shown in \tabref{tab: strategy}, we conduct ablation studies to explore key design choices in \methodname{}, 
including target VLMs for aligning representations, region cropping, 
and optimization strategies with a focus on GPU memory usage and time overhead.
Using a self-distillation strategy~\cite{wu2023clipself} as the baseline,
we observe a significant performance drop when replacing the target model with ViT-L/14 CLIP from CLIPSelf, 
due to disruption in visual-language alignment.
Replacing feature distillation with logit distillation improves performance but decreases training efficiency 
due to repeated feature extraction while introducing foreground bias issues.
In contrast, \methodname{}  leverages a more efficient feature-cropping strategy and a decoupled alignment framework to achieve superior performance while substantially reducing both training time and GPU memory usage.

\begin{table}[t]
    \centering
    \tablestyle{4.5pt}{1.0}
    \centering
    \small
    \begin{tabular}{c c cc c c c }
    \toprule
    \multicolumn{2}{c}{P-Thing}  &\multicolumn{2}{c}{P-Stuff} \\
    \cmidrule(lr){1-2}
    \cmidrule(lr){3-4}
    \textit{Thing}&\textit{Stuff}&\textit{Thing}&\textit{Stuff}&\textbf{Boxes}  &\textbf{Masks-T}&\textbf{Masks-S}\\
    \midrule
    \multicolumn{1}{c}{\cmark} &\cmark  &\cmark &\cmark & \textbf{74.3}& 70.9& 42.6  \\
    \multicolumn{1}{c}{\cmark} &\xmark  &\xmark &\cmark & 74.2& 70.8& 42.0  \\
    \multicolumn{1}{c}{\cmark} &\cmark  &\xmark &\cmark & 74.1& 70.9& 41.3  \\
    \rowcolor[HTML]{EFEFEF}\multicolumn{1}{c}{\cmark} &\xmark  &\cmark &\cmark & 73.4& \textbf{71.0}& \textbf{45.6} \\
    \bottomrule
    \end{tabular}
  \vspace{-5pt}
  \caption{
  {Ablation study on decoupled alignment}.
  \cmark denotes that a region is separated from the categories in this set.
  }
  \label{tab: decouple}
\vspace{-12pt}
\end{table}

\myPara{Ablation study on decoupled alignment.}
\tabref{tab: decouple} analyzes the effect of the decoupled alignment strategy on \methodname{}. 
The retrieved category $c^k$ can be categorized into \textit{Thing} and \textit{Stuff}, 
referred to as `P-Thing' and `P-Stuff', respectively.
\methodname{} leverages $c^k$ for regions to selectively contrast against irrelevant \textit{Thing} and \textit{Stuff} categories.
When regions contrast against all categories,
the model can better identify the foreground regions but struggles with distinguishing the background (1st row).
In the fully decoupled setting, where each region contrasts against its own set of categories,
performance further degrades (2nd row).
The most pronounced foreground bias occurs when P-Thing contrasts against \textit{Stuff} but P-Stuff does not contrast against \textit{Thing} (3rd row).
In \methodname{}, we adopt a balanced strategy where P-Stuff contrasts against both \textit{Thing} and \textit{Stuff}, 
while P-Thing does not contrast against \textit{Stuff},
resulting in improved overall performance by mitigating bias (4th row).

\begin{figure*}[t]
  \centering
  \begin{overpic}[width=0.98\textwidth]{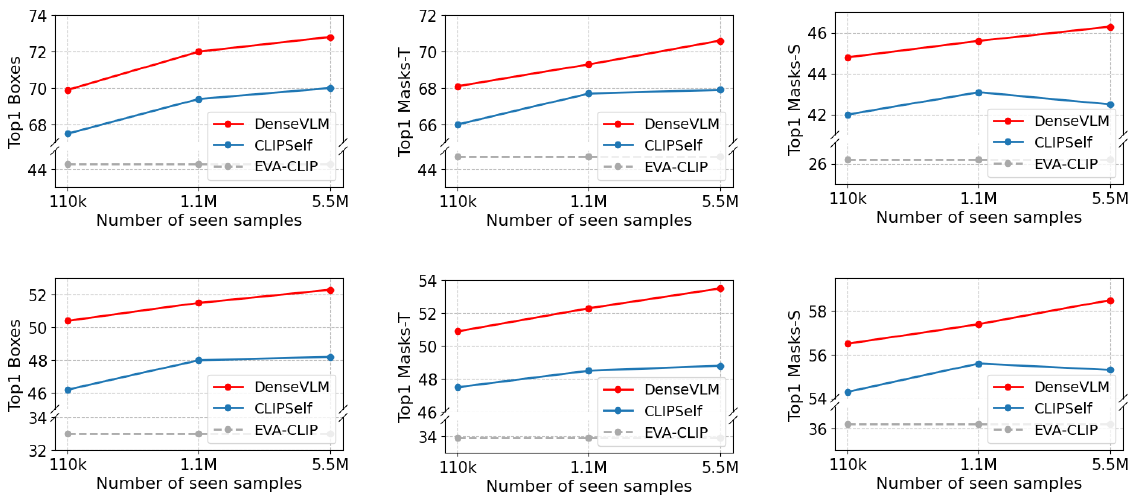}
  \put(20,21.5){(a) Top1 mean accuracy of models on COCO dense prediction benchmark.}
  \put(20,-1.8){(b) Top1 mean accuracy of models on ADE20K dense prediction benchmark.}

  \end{overpic}
  \vspace{7pt}
  \caption{
  {Zero-shot comparisons of models pre-trained on datasets with three different scales.}
  We select three training sets from the SA-1B dataset~\cite{kirillov2023segment}: 100K, 1.1M, and 5.5M seen samples
  and perform the zero-shot evaluation on the COCO and ADE20K benchmarks.
  }
  \label{fig: data_scaling}
  \vspace{-10pt}
\end{figure*}

\myPara{Ablation study on different category sets.}
\tabref{tab: categories} presents the effects of different category sets used for retrieval and alignment performance. 
When using the category set from dataset annotations,
we observe that as the number of categories increases, performance consistently improves.
Adding only the background category helps the model distinguish foreground differences, 
as seen in rows 1 and 2, 
and incorporating more foreground categories leads to performance gains,
as seen in rows 3 and 4.
These improvements can be attributed to the enhanced representational capacity 
afforded by a broader category set.

To address the limitations of category set annotations, 
we generate category sets using the generative model NXTP~\cite{Yue_2024_CVPR} based on images 
and use DeepSeek-R1~\cite{guo2025deepseek} to classify foreground and background categories. 
As shown in \tabref{tab: categories}, our model still achieves strong performance.

\begin{table}[t]
    \centering
    \tablestyle{6.pt}{1.0}
    \centering
    \small
    \begin{tabular}{l cc cc cc}
    \toprule
      &\multicolumn{2}{c}{\textbf{Boxes}}  &\multicolumn{2}{c}{\textbf{Masks-T}}&\multicolumn{2}{c}{\textbf{Masks-S}} \\
    \multirow{1}{*}{Categories} &Top1&Top5&Top1&Top5& Top1&Top5\\
    \midrule
    \multicolumn{7}{l}{\textit{\textbf{The category set is from the dataset annotation.}}}\\
    133 (80)& 71.1& 88.5& 68.7& 83.0& 44.7& 75.2 \\
    171 (80)& 72.3& 89.8& 69.4& 85.8& 44.2& 76.0 \\
    273 (160)& 72.3& 89.9& 70.1& 84.4& 44.9& 76.4 \\
    316 (204)& \textbf{73.4}& \textbf{90.5}& \textbf{71.0}& \textbf{84.8}& \textbf{45.6}& \textbf{77.8}\\
    \hline
    \multicolumn{7}{l}{\textit{\textbf{The category set is generated by NXTP based on the images.}}}\\
    210 (133)& \textbf{72.6}& \textbf{90.2}& \textbf{70.4}& {84.6}& 41.5& 75.7 \\
    794 (484)& 72.5& 90.4& 70.3& \textbf{84.8}& \textbf{44.1}& \textbf{76.2} \\

    \bottomrule
    \end{tabular}
  \vspace{-5pt}
  \caption{
  {Ablation study on different category sets}.
  The numbers in parentheses denote the number of foreground thing classes.
  }
  \label{tab: categories}
\vspace{-15pt}
\end{table}

\myPara{Ablation study on different P-VLMs.}
\tabref{tab: pvlm} shows that using ViT-B/16 as the retrieval model, 
our method performs well in mask classification, 
outperforming the P-VLM (ViT-B/16 with CLIPSelf).
This improvement effectively mitigates foreground bias by enhancing the distinction between foreground and background regions.
Using ViT-L/14 further improves performance across all metrics.

\subsection{Zero-shot comparisons at different data scales}
To investigate the effect of data scale on VLM performance, 
we select three training sets of varying sizes from the SA-1B~\cite{kirillov2023segment} dataset: 100K, 1.1M, and 5.5M seen samples. 
We train our approach and the competing approach CLIPSelf on these subsets using 8$\times$A40 GPUs and batch size of 48 per GPU.
%
The resulting accuracy lines, presented in~\figref{fig: data_scaling}, 
reflect the performance of each model on the COCO and ADE20K benchmarks for open-vocabulary dense prediction tasks.
As the size of the training set increases, the performance improvement of CLIPSelf slows down,
while \methodname{} continues to show consistent improvement, demonstrating its promising scalability.
The ability of our \methodname{} to rely on unlabeled data enables it to scale effectively to larger datasets.
%

\begin{table}[t]
    \centering
    \tablestyle{6.pt}{1.0}
    \centering
    \small
    \begin{tabular}{l cc cc cc}
    \toprule
      &\multicolumn{2}{c}{\textbf{Boxes}}  &\multicolumn{2}{c}{\textbf{Masks-T}}&\multicolumn{2}{c}{\textbf{Masks-S}} \\
    \multirow{1}{*}{P-VLM } &Top1&Top5&Top1&Top5& Top1&Top5\\
    \midrule
    ViT-B/16& 70.0& 88.6& 68.0& 83.0& 43.0& 75.3 \\
    ViT-L/14 & \textbf{73.4}& \textbf{90.5}& \textbf{71.0}& \textbf{84.8}& \textbf{45.6}& \textbf{77.8}\\
    \bottomrule
    \end{tabular}
  \vspace{-5pt}
  \caption{
  Ablation study on different P-VLMs utilized for retrieving categories for unlabeled regions.
  }
  \label{tab: pvlm}
\vspace{-10pt}
\end{table}

\begin{table}[t]
  \centering
  \tablestyle{5.5pt}{1.0}
  \centering
  \small
  \begin{tabular}{l c c c c c}
  \toprule
  \multirow{1}{*}{Backbones}&\multirow{1}{*}{VLMs}& \multicolumn{1}{c}{\textbf{Boxes}}& \multicolumn{1}{c}{\textbf{Masks-T}}&\multicolumn{1}{c}{\textbf{Masks-S}} \\
  \midrule

   ViT-L/14 &CLIPSelf & 75.2& 73.1& 44.5  \\
   \hline
   ViT-B/16 &OpenCLIP& 49.8& 51.9& 29.2  \\
   ViT-B/16* &CLIPSelf & 67.6& 64.4& 44.5  \\
   \cellcolor[HTML]{EFEFEF}ViT-B/16* &\cellcolor[HTML]{EFEFEF}\methodname{} & \cellcolor[HTML]{EFEFEF}\textbf{71.9}& \cellcolor[HTML]{EFEFEF}\textbf{70.0}& \cellcolor[HTML]{EFEFEF}\textbf{47.8}  \\

   \hline
   ViT-B/16&EVA-CLIP& 44.3& 44.7& 26.2   \\
   ViT-B/16 &CLIPSelf & 69.1& 66.7& 41.7  \\
   \cellcolor[HTML]{EFEFEF}ViT-B/16 &\cellcolor[HTML]{EFEFEF}\methodname{} & \cellcolor[HTML]{EFEFEF}\textbf{73.4}& \cellcolor[HTML]{EFEFEF}\textbf{71.0}& \cellcolor[HTML]{EFEFEF}\textbf{45.6} \\

   \hline
   R50x4&OpenCLIP  &59.2& 50.5& 39.1 \\
   R50x4 &CLIPSelf &59.1& 49.9& 37.2  \\
   \cellcolor[HTML]{EFEFEF}R50x4 &\cellcolor[HTML]{EFEFEF}\methodname{} & \cellcolor[HTML]{EFEFEF}\textbf{65.6}& \cellcolor[HTML]{EFEFEF}\textbf{55.2}& \cellcolor[HTML]{EFEFEF}\textbf{40.9} \\

   \hline
   ConvNeXt-B&OpenCLIP  & 57.5& 48.0& 31.1 \\
   ConvNeXt-B &CLIPSelf & 62.6& 57.6& 41.6  \\
   \cellcolor[HTML]{EFEFEF}ConvNeXt-B &\cellcolor[HTML]{EFEFEF}\methodname{}& \cellcolor[HTML]{EFEFEF}\textbf{67.1}& \cellcolor[HTML]{EFEFEF}\textbf{63.4}& \cellcolor[HTML]{EFEFEF}\textbf{43.5}  \\

  \bottomrule
  \end{tabular}
  \vspace{-5pt}
\caption{
{Results and comparisons of different backbones}.
* indicates the model initialized by OpenCLIP~\cite{openclip}.
}
\label{tab: other_backbones}
\end{table}


\begin{table}[t]
  \centering
  \tablestyle{6.pt}{1.0}
  \small
  \begin{tabular}{l llll }
  \toprule
  Method&\textbf{A-150} & \textbf{A-847}& \textbf{PC-459}\\
  \midrule
  PACL~\cite{mukhoti2023open}& 31.4& -& -\\
  OVSeg~\cite{ovseg} & 24.8& 7.1& 11.0 \\
  MAFT~\cite{jiao2023learning} & 29.1& 10.1& 12.6\\
  SED~\cite{xie2023sed} & 31.6& 11.4& 18.6\\
  SCAN~\cite{Liu_2024_CVPR}& 30.8& 10.8& 13.2\\
  CAT-Seg+CLIPSelf~\cite{wu2023clipself}&29.7&10.1&-\\
  CAT-Seg+FineCLIP~\cite{jing2025fineclip}&32.4&12.2&-\\
  \hline
  SAN~\cite{san}& 27.4 & 10.0 & 13.0 \\
  \cellcolor[HTML]{EFEFEF}SAN+\methodname{}& \cellcolor[HTML]{EFEFEF}\textbf{29.5}$_{\textcolor{red}{+2.1}}$& \cellcolor[HTML]{EFEFEF}\textbf{10.4}$_{\textcolor{red}{+0.4}}$ & \cellcolor[HTML]{EFEFEF}\textbf{15.6}$_{\textcolor{red}{+2.6}}$    \\
  \hline
  CAT-Seg~\cite{catseg} & 31.4&11.7&18.4\\
  \cellcolor[HTML]{EFEFEF}CAT-Seg+\methodname{} & \cellcolor[HTML]{EFEFEF}\textbf{34.1}$_{\textcolor{red}{+2.7}}$& \cellcolor[HTML]{EFEFEF}\textbf{12.2}$_{\textcolor{red}{+0.5}}$&\cellcolor[HTML]{EFEFEF}\textbf{18.7}$_{\textcolor{red}{+0.3}}$\\
  \bottomrule
  \end{tabular}
  \vspace{-5pt}
  \caption{
  {Results on open-vocabulary semantic segmentation}.
  %
  }
  \label{tab: ovs}
  \vspace{-15pt}
\end{table}

\subsection{Extending to different backbones}
To demonstrate the generality of our method, we extend it to several backbones, 
including ViT-B/16, R50x4, and ConvNeXt-B from OpenCLIP~\cite{openclip}, as well as ViT-B/16 from EVA-CLIP~\cite{sun2023eva}.
From~\tabref{tab: other_backbones}, we observe that CLIPSelf~\cite{wu2023clipself} is more effective with ViT-based architectures 
but is less effective with CNN-based architectures, 
particularly showing degraded performance with R50x4 compared to the baseline.
In contrast, our approach yields consistent performance gains across all architectures, 
thereby broadening its applicability and potential.
Furthermore, \methodname{} narrows the performance gap between powerful models like ViT-L/14 fine-tuned by CLIPSelf and lightweight models, 
with some lightweight models even outperforming the more powerful ones on the Masks-S metric.

\subsection{Application to open-vocabulary dense tasks}

\myPara{Experiment settings.}
To evaluate the performance of the proposed \methodname{} in downstream tasks, 
we use \methodname{} as the backbone for open-vocabulary dense prediction tasks, 
including object detection and segmentation.
To ensure fairness, our \methodname{} models are trained on the COCO train2017 split with an input resolution of 512$\times$512.

\begin{table}[t]
  \centering
  \tablestyle{2pt}{1.0}
  \vspace{-5pt}
  \begin{minipage}[t]{0.48\textwidth}
  \small
  \subcaption{OV-COCO benchmark}
  \begin{tabular}{l l lll}
      \toprule
      Method& Backbone& \textbf{AP}$_{50}^{\mathrm{novel}}$& \textbf{AP}$_{50}^{\mathrm{base}}$&\textbf{AP}$_{50}$\\
      \midrule
      VLDet~\cite{lin2022learning}& RN50& 32.0& 50.6& 45.8\\
      F-VLM~\cite{kuo2022f}&  RN50& 28.0& -& 39.6\\
      BARON-Cap~\cite{Wu_2023_CVPR}&  RN50& 33.1& 54.8& 49.1\\
      CORA~\cite{CORA}& RN50& 35.1& 35.5& 35.4 \\
      RO-ViT~\cite{Kim_2023_CVPR}& ViT-B/16& 30.2&- & 41.5 \\
      RO-ViT~\cite{Kim_2023_CVPR}& ViT-L/16& 33.0&- & 47.7 \\
     F-ViT+CLIPSelf~\cite{wu2023clipself} & ViT-B/16& 25.4& 40.9& 36.8\\
     F-ViT+FineCLIP~\cite{jing2025fineclip} & ViT-B/16& 29.8& 45.9& 41.7\\
      \hline
      F-ViT~\cite{wu2023clipself} & ViT-B/16& 17.5& 41.0& 34.9\\
      \cellcolor[HTML]{EFEFEF}F-ViT+\methodname & \cellcolor[HTML]{EFEFEF}ViT-B/16&\cellcolor[HTML]{EFEFEF}\textbf{33.1}$_{\textcolor{red}{+15.6}}$&\cellcolor[HTML]{EFEFEF}\textbf{52.5}$_{\textcolor{red}{+11.5}}$&\cellcolor[HTML]{EFEFEF}\textbf{47.4}$_{\textcolor{red}{+12.5}}$\\
      \bottomrule
  \end{tabular}
  \vspace{5pt}
  \end{minipage}
  \begin{minipage}[t]{0.48\textwidth}
      \small
      \subcaption{OV-LVIS benchmark}
      \begin{tabular}{l llll }
          \toprule
          Method& Backbone&\quad \textbf{mAP}$_{r}$& \textbf{mAP}$_{c}$&\textbf{mAP}\\
          \midrule
          VLDet~\cite{lin2022learning}& RN50&\quad 21.7& 29.8& 30.1\\
          BARON-Cap~\cite{Wu_2023_CVPR}&  RN50&\quad 22.6& 27.6& 27.6\\
          F-VLM~\cite{kuo2022f}&  RN50&\quad 18.6& -& 24.2\\
          COR~\cite{CORA}& RN50x4&\quad 22.2& -& - \\
          RO-ViT~\cite{Kim_2023_CVPR}& ViT-B/16&\quad 28.0&- & 30.2 \\
         F-ViT+CLIPSelf~\cite{wu2023clipself} & ViT-B/16& \quad10.6& 7.6& 9.3\\
     F-ViT+FineCLIP~\cite{jing2025fineclip} & ViT-B/16& \quad 10.4& 8.0& 9.5\\
          \hline
          F-ViT~\cite{wu2023clipself} & ViT-B/16&\quad 11.5& 12.3& 15.4\\
          \cellcolor[HTML]{EFEFEF}F-ViT+\methodname & \cellcolor[HTML]{EFEFEF}ViT-B/16&\cellcolor[HTML]{EFEFEF}\quad \textbf{23.9}$_{\textcolor{red}{+12.4}}$&\cellcolor[HTML]{EFEFEF}\textbf{18.4}$_{\textcolor{red}{+6.1}}$&\cellcolor[HTML]{EFEFEF}\textbf{21.4}$_{\textcolor{red}{+6.0}}$\\
          \bottomrule
      \end{tabular}
  \end{minipage}
  \vspace{-5pt}
  \caption{
  {Results on open-vocabulary object detection}. 
  %
  }
  \label{tab: ovd}
  \vspace{-11pt}
\end{table}

\myPara{Open-vocabulary semantic segmentation.}
\label{exp: ovs}
We apply \methodname{} models initialized with OpenAI CLIP~\cite{CLIP} to open-vocabulary semantic segmentation, 
including SAN~\cite{san} with a frozen backbone and Cat-Seg~\cite{catseg} with a fine-tuned backbone. 
The models are trained on COCO-Stuff~\cite{caesar2018coco} and evaluated on the ADE20K~\cite{ade20k} (ADE-150 and ADE-847) 
and PASCAL Context~\cite{pascal_context} (PC-459) datasets, using the mean Intersection-over-Union (mIoU) metric.
As shown in~\tabref{tab: ovs}, \methodname{} consistently improves performance across all evaluation datasets, 
further enhancing the state-of-the-art performance.

\myPara{Open-vocabulary object detection.}
\label{exp: ovd}
Building on previous work~\cite{wu2023clipself}, we construct open-vocabulary object detectors 
using the F-ViT architecture, 
a two-stage detector built on frozen ViTs from EVA-CLIP~\cite{sun2023eva}.
%
%
%
As shown in~\tabref{tab: ovd}, we evaluate performance on the OV-COCO~\cite{chen2015microsoft} benchmark by reporting box AP at IoU 0.5 for base, novel, and overall categories (AP$_{50}^{\mathrm{novel}}$, AP$_{50}^{\mathrm{base}}$, AP$_{50}$), 
and on OV-LVIS~\cite{Gupta_2019_CVPR} benchmark by mean AP for rare (mAP$_{r}$), common (mAP$_{c}$), and all categories (mAP).
Replacing the frozen EVA-CLIP ViTs with \methodname{} models leads to clear performance improvements on both benchmarks, 
achieving competitive performance relative to previous methods.
%

\section{Conclusions}
In this paper, we present \methodname{}, 
a framework designed to mitigate foreground bias in region-level vision-language alignment.
\methodname{} can directly replace the original VLMs in open-vocabulary object detection and image segmentation methods, 
demonstrating consistently clear performance improvements over the baseline models.
Furthermore, we validate \methodname{}'s promising scaling ability by exploring efficient region retrieval and decoupled alignment, 
successfully implementing \methodname{} to the training data from the SA-1B dataset. 
Overall, \methodname{} offers a generalizable solution for improving dense representations in vision-language models
across various backbones, which is essential for open-vocabulary dense prediction tasks.

\section*{Acknowledgments}
This work was partially funded by NSFC (No. 62225604, 62276145) and the Science and Technology Support Program of Tianjin, China (No. 23JCZDJC01050).
This work was also supported by the Supercomputing Center of Nankai University.

{
\small
\bibliographystyle{ieeenat_fullname}
\bibliography{refs}
}


\maketitlesupplementary

We provide an overview of the supplementary materials to ensure a clear and comprehensive understanding.

\begin{itemize}
\item In~\secref{append: limitation}, we detail the limitations and broader impact of \methodname.
\item In~\secref{append: technical_details}, we present the training details.
\item In~\secref{append: experiments}, we offer supplementary experiments on 
input image sizes, region proposals, backbones, category sets and threshold of region denoising for training VLMs.
\item In~\secref{append: visualization}, we show visualizations, including confusion matrices and croppping feayures predictions, to demonstrate foreground bias. .
\item In~\secref{append: dataset}, we present the dataset information for training and evaluation.
\end{itemize}

\section{Limitations and broader impact}
\label{append: limitation}
\noindent\textbf{Limitations:}
Our aim is to develop a region-language alignment model that effectively integrates local visual and semantic features, thereby improving open-vocabulary dense prediction performance.
Compared to previous pre-trained Vision-Language Models (VLMs)~\cite{openclip,sun2023eva,zhong2022regionclip,wu2023clipself}, 
our proposed \methodname{} achieves superior results and significantly improves downstream task performance. 
We believe \methodname{} has even greater potential. 
1) Scalability. \methodname{} is designed within an efficient, 
unsupervised region-language alignment framework, making it adaptable to various datasets. 
However, computational resource limitations have restricted our ability to scale to larger datasets. 
2) Model capacity. We employ the ViT-L/14 model from CLIPSelf~\cite{wu2023clipself} as a powerful Pre-trained VLM (P-VLM).
Utilizing more robust VLMs can yield better performance, 
and transferring their rich semantic knowledge to training models is a promising direction. 
3) Fine-grained Semantic. We categorize objects into broad \textit{thing} and \textit{stuff} classes. 
Fine-grained semantic segmentation and decoupled alignment would enhance the model's ability to distinguish between similar categories.
We plan to explore these avenues in our future research.

\noindent\textbf{Broader impact:}
\methodname~exhibits notable potential for open-vocabulary dense predictions within scenes, 
which can enhance various applications such as robotics and environmental monitoring.
By enabling systems to recognize and interpret a wide range of objects and contexts without prior training on specific categories, 
\methodname{} facilitates more adaptive and versatile applications.
Given its broad applicability and non-specialized nature, our method is designed to support a variety of 
technical advancements without directly addressing specific societal challenges.

\begin{table}[t]
    \small
    \centering
    \vspace{5pt}
    \tablestyle{13pt}{1.1}
    \begin{tabular}{l c} 
        \toprule
      item & value \\ 
          \midrule
      image size    & 512 $\times$ 512\\
      optimizer & AdamW~\cite{adamw} \\
      learning rate & 0.0001 \\
      $\beta_1$ & 0.9\\
      $\beta_2$ & 0.98\\
      weight decay & 0.1 \\
      batch size (per card)& 48 \\
      warmup steps~\cite{goyal2018accurate} & 1000 \\
      epochs & 6 \\
      learning rate scheduler & cosine decay~\cite{loshchilov2017sgdr} \\
      number of GPUs& 4 \\
      automatic mixed precision${}^1$ & True \\
          \bottomrule
    \end{tabular}
    \caption{Training details of \methodname.}
    \label{tab: training_details}
    \vspace{-5pt}
\end{table}

\renewcommand{\thefootnote}{\arabic{footnote}}

\footnotetext[1]{https://developer.nvidia.com/automatic-mixed-precision}

\section{Training details}
\label{append: technical_details}
We train all models on NVIDIA A40 GPUs to ensure a fair comparison across experiments.
The detailed configuration is provided in Table~\ref{tab: training_details}. 
For the SA-1B dataset~\cite{kirillov2023segment}, we use 8$\times$A40 GPUs to ensure efficient and scalable training.

For open-vocabulary segmentation, 
we train the models such as SAN~\cite{san} and CAT-Seg~\cite{catseg} on the COCO-Stuff~\cite{caesar2018coco} dataset for 80$k$ iterations.
For open-vocabulary detection,
models are trained for 3 epochs on the OV-COCO~\cite{chen2015microsoft} benchmark and 48 epochs on OV-LVIS~\cite{Gupta_2019_CVPR} benchmark.

\begin{table*}[t]
    \centering
    \tablestyle{6.5pt}{1.05}
    \centering
    \small
    \begin{tabular}{c c  cc cc cc cc cc cc}
    \toprule
    && \multicolumn{6}{c}{COCO}& \multicolumn{6}{c}{ADE20K}\\
    \cmidrule(lr){3-8}
    \cmidrule(lr){9-14}
   Input& GPU Memory& \multicolumn{2}{c}{\textbf{Boxes}}  &\multicolumn{2}{c}{\textbf{Masks-T}}&\multicolumn{2}{c}{\textbf{Masks-S}} & \multicolumn{2}{c}{\textbf{Boxes}}  &\multicolumn{2}{c}{\textbf{Masks-T}}&\multicolumn{2}{c}{\textbf{Masks-S}} \\
    \multirow{1}{*}{Image Size}&(per card)&Top1&Top5&Top1&Top5& Top1&Top5&Top1&Top5&Top1&Top5& Top1&Top5 \\
    \midrule
    224& 9G &{60.1}& {79.9}& {49.4}& {62.4}& {35.3}& {64.2}& {40.0}& {70.0}& {36.3}& {56.4}& {50.3}& {77.0} \\ 
    320& 11G &{66.2}& {85.4}& {59.2}& {73.0}& {41.0}& {71.2}& {45.6}& {76.0}& {44.0}& {67.6}& {54.3}& {81.7} \\ 
    512& 16G &{73.4}& {90.5}& {71.0}& {84.8}& {45.6}& {77.8}& {51.3}& {82.2}& {52.1}& {78.0}& {57.8}& {85.5} \\ 
    768& 27G  &{74.4}& {91.3}& {75.4}& {90.1}& {45.5}& {79.0}& {52.7}& {82.9}& {55.4}& {82.6}& {58.2}& {86.6} \\ 
    1024& 39G &\textbf{76.6}& \textbf{93.1}& \textbf{78.7}& \textbf{93.6}& \textbf{46.5}& \textbf{79.8}& \textbf{53.2}& \textbf{83.6}& \textbf{56.8}& \textbf{83.2}& \textbf{58.6}& \textbf{86.8} \\ 
    \bottomrule
    \end{tabular}
    \vspace{-5pt}
  \caption{
  {Ablation study on input image sizes}.
  We report the Top1 and Top5 mean accuracy on classifying boxes and panoptic masks
  on COCO panoptic and ADE20K panoptic benchmarks.
  The GPU memory usage corresponds to a batch size of 12 on A40 GPU.
  }
  \label{tab: image_size}
  \vspace{-5pt}
\end{table*}

\begin{table*}[t]
    \centering
    \tablestyle{6.pt}{1.1}
    \centering
    \small
    \begin{tabular}{c c  cc cc cc cc cc cc}
    \toprule
    && \multicolumn{6}{c}{COCO}& \multicolumn{6}{c}{ADE20K}\\
    \cmidrule(lr){3-8}
    \cmidrule(lr){9-14}
    & & \multicolumn{2}{c}{\textbf{Boxes}}  &\multicolumn{2}{c}{\textbf{Masks-T}}&\multicolumn{2}{c}{\textbf{Masks-S}} & \multicolumn{2}{c}{\textbf{Boxes}}  &\multicolumn{2}{c}{\textbf{Masks-T}}&\multicolumn{2}{c}{\textbf{Masks-S}} \\
    \multirow{1}{*}{Method}&Region Proposals&Top1&Top5&Top1&Top5& Top1&Top5&Top1&Top5&Top1&Top5& Top1&Top5 \\
    \midrule
    CLIPSelf& \xmark&69.1& 88.2& 66.7& 83.0& 41.7& 75.2& 48.1& 77.7& 47.5& 74.2& 53.7& 82.8\\
    CLIPSelf& \cmark&{70.2}& {89.2}& {68.1}& {83.5}& {35.7}& {71.8}& {49.8}& {79.7}& {51.5}& {76.0}& {50.9}& {80.7} \\ 
    \hline
    \methodname&\xmark &{73.4}& {90.5}& {71.0}& {84.8}& {45.6}& {77.8}& {51.3}& {82.2}& {52.1}& {78.0}& {57.8}& {85.5} \\ 
    \methodname&\cmark &{74.4}& {91.3}& {75.4}& {90.1}& {45.9}& {79.0}& {52.7}& {82.9}& {55.4}& {82.6}& {58.2}& {86.6} \\ 
    \bottomrule
    \end{tabular}
    \vspace{-5pt}
  \caption{
  {Ablation study on using region proposals}.
  We report the Top1 and Top5 mean accuracy on classifying boxes and panoptic masks (thing and stuff)
  on COCO panoptic and ADE20K panoptic benchmarks.
  }
  \label{tab: proposals}
\end{table*}

\section{Additional experiments}
\label{append: experiments}

\myPara{Ablation study on input image sizes.}
To evaluate the effect of input image size on \methodname{}, we conduct experiments 
with distinct resolutions: 224, 320, 512, 768 and 1024 pixels, for both training and inference. 
As shown in~\tabref{tab: image_size}, 
model's performance on the region classification task improves as image resolution increases from 224 to 1024 pixels. 
This enhancement can be attributed to the greater detail captured at higher resolutions. 
However, this improvement comes a significant increase in GPU memory usage.
Considering the trade-off between computational resources and model performance, 
we resize the images to 512$\times$512 pixels to achieve an optimal balance.

\begin{table}[t]
    \centering
    \tablestyle{8pt}{1.0}
    \centering
    \small
    \begin{tabular}{l cc cc cc}
    \toprule
      &\multicolumn{2}{c}{\textbf{Boxes}}  &\multicolumn{2}{c}{\textbf{Masks-T}}&\multicolumn{2}{c}{\textbf{Masks-S}} \\
    \multirow{1}{*}{$\theta$}&Top1&Top5&Top1&Top5& Top1&Top5\\
    \midrule
    0.0& 72.1& 89.6& 68.2& 84.3& 43.6& 76.1 \\
    0.1& 72.7& 90.2& 69.1& 84.3& 44.6& 77.2 \\
    0.2& 73.1& 90.4& 69.7& 84.6& 45.1& 77.7 \\
    0.3&\textbf{73.4}& \textbf{90.5}& \textbf{71.0}& \textbf{84.8}& \textbf{45.6}& \textbf{77.8} \\
    0.4& 73.2& 90.2& 70.2& 84.3& 45.2& 77.5 \\
    0.5& 73.1& 90.0& 70.0& 84.3& 45.0& 77.1 \\
    0.6& 73.1& 89.9& 69.6& 84.0& 44.6& 76.3 \\
    \bottomrule
    \end{tabular}
  \vspace{-5pt}
  \caption{
  {Ablation study on threshold of $\theta$ in region denoising}.
  }
  \label{tab: threshold}
\vspace{-5pt}
\end{table}

\myPara{Ablation study on using region proposals.}
Following RegionCLIP~\cite{zhong2022regionclip} for fine-tuning VLMs with pseudo-labelled region-text pairs, 
we compare our approach to CLIPSelf~\cite{wu2023clipself} in utilizing these pairs.
As shown in~\tabref{tab: proposals}, 
CLIPSelf substitutes random image crops with pseudo region-text pairs, 
resulting in an enhanced recognition for foreground objects while concurrently observing a reduction in the accuracy of background identification.
In contrast, our proposed \methodname{} achieves a notable improvement in the recognition accuracy of foreground objects 
while also improving the identification of background stuff.

\myPara{Ablation study on the threshold $\theta$.}
We perform an ablation experiment to assess the impact of varying threshold $\theta$ values of region denoising.
As shown in~\tabref{tab: threshold}, the model performs the worst when $\theta=0$.
When $\theta$ is set lower, the Top-5 accuracy increases, but results in suboptimal performance. 
This may be due to low-confidence categories causing alignment confusion for the model.
Conversely, setting $\theta$ too high filters out too many local images, decreasing performance. 
By defaul, we select $\theta = 0.3$ for \methodname{}.

\myPara{Ablation study on various backbones.}
\methodname{} exhibits adaptability to diverse backbones. 
As shown in~\tabref{tab: openai_backbones}, our models achieve consistent superiority over prior approaches~\cite{openclip,wu2023clipself} across all dense prediction tasks.
Particularly, the ViT-B/16-based \methodname{} performs comparably to the ViT-L/14-based CLIPSelf~\cite{wu2023clipself}. 
Utilizing ViT-L/14 with a large number of parameters as initialization, 
\methodname{} achieves clearly enhancements across all evaluated metrics, 
thereby facilitating superior performance in dense prediction tasks.

\begin{table*}[t]
    \centering
    \tablestyle{7.pt}{1.05}
    \centering
    \small
    \begin{tabular}{l l  cc cc cc cc cc cc}
    \toprule
    && \multicolumn{6}{c}{COCO}& \multicolumn{6}{c}{ADE20K}\\
    \cmidrule(lr){3-8}
    \cmidrule(lr){9-14}
    & & \multicolumn{2}{c}{\textbf{Boxes}}  &\multicolumn{2}{c}{\textbf{Masks-T}}&\multicolumn{2}{c}{\textbf{Masks-S}} & \multicolumn{2}{c}{\textbf{Boxes}}  &\multicolumn{2}{c}{\textbf{Masks-T}}&\multicolumn{2}{c}{\textbf{Masks-S}} \\
    \multirow{1}{*}{backbones}&VLMs&Top1&Top5&Top1&Top5& Top1&Top5&Top1&Top5&Top1&Top5& Top1&Top5 \\
    \midrule
    ViT-B/16& OpenCLIP&49.8& 74.3& 51.9& 72.2& 29.2& 54.9& 28.4& 54.1& 29.6& 53.4& 37.9& 66.6\\
    ViT-B/16& CLIPSelf & {67.6}& {87.8}& {64.4}& {81.2}& {44.5}& {77.1}&43.4& {76.0}& {44.0}& {71.1}& {50.7}& {82.1} \\ 
    ViT-B/16& \methodname{}*&{71.9}& {90.2}& {70.0}& {84.3}& {47.8}& {79.4}& {48.5}& {79.2}& {49.0}& {75.2}& {55.1}& {85.2} \\ 
    ViT-B/16& \methodname{} &{73.4}& {90.5}& {71.0}& {84.8}& {45.6}& {77.8}& {51.3}& {82.2}& {52.1}& {78.0}& {57.8}& {85.5} \\ 
    \hline
    ViT-L/14& OpenCLIP&21.2& 45.3& 26.6& 48.9& 11.2& 27.2& 48.1& 11.9& 34.1& 13.9& 11.1& 32.4\\
    ViT-L/14& CLIPSelf&68.3 &{90.1}& {67.1}& {84.5}& {37.7}& {71.3}& {47.1}& {77.5}& {47.7}& {74.4}& {48.9}&82.3 \\ 
    ViT-L/14& \methodname{}*&{76.2}& {92.9}& {73.3}& {87.3}& {47.4}& {79.1}& {54.0}& {84.1}& {54.2}& {79.9}& {57.8}& {85.9} \\ 
    ViT-L/14& \methodname{}&{75.2}& {91.8}& {73.3}& {87.1}& {45.5}& {78.1}& {54.5}& {85.0}& {55.6}& {82.1}& {58.1}& {86.4} \\ 
    \bottomrule
    \end{tabular}
    \vspace{-5pt}
  \caption{
  {Ablation study on various backbones}.
  We report the Top1 and Top5 mean accuracy on classifying boxes and panoptic masks (thing and stuff)
  on COCO panoptic and ADE20K panoptic benchmarks.
  * indicates the model initialized by OpenCLIP~\cite{openclip}.
  }
  \label{tab: openai_backbones}
\end{table*}

\begin{table*}[t]
    \centering
    \tablestyle{9pt}{1.0}
    \centering
    \small
    \begin{tabular}{l cc cc cc cc cc cc}
    \toprule
    &\multicolumn{6}{c}{COCO}& \multicolumn{6}{c}{ADE20K}\\
    \cmidrule(lr){2-7}
    \cmidrule(lr){8-13}
    & \multicolumn{2}{c}{\textbf{Boxes}}  &\multicolumn{2}{c}{\textbf{Masks-T}}&\multicolumn{2}{c}{\textbf{Masks-S}} & \multicolumn{2}{c}{\textbf{Boxes}}  &\multicolumn{2}{c}{\textbf{Masks-T}}&\multicolumn{2}{c}{\textbf{Masks-S}} \\
    \multirow{1}{*}{Categories}&Top1&Top5&Top1&Top5& Top1&Top5&Top1&Top5&Top1&Top5& Top1&Top5 \\
    \midrule
    133 (80)&71.1& 88.5& 68.7& 83.0& 44.7& 75.2& 49.4& 79.0& 48.5& 74.2& 54.7& 82.8\\
    171 (80)&72.3& 89.8& 69.4& 85.8& 44.2& 76.0& {49.8}& {79.7}& {48.9}& {75.1}& {55.1}& {82.4} \\ 
    273 (160) &72.3& 89.9& 70.1& 84.4& 44.9& 76.4& {51.0}& {81.8}& {49.3}& {76.5}& {57.0}& {84.0} \\ 
    316 (204) &\textbf{73.4}& \textbf{90.5}& \textbf{71.0}& \textbf{84.8}& \textbf{45.6}& \textbf{77.8}& \textbf{51.3}& \textbf{82.2}& \textbf{52.1}& \textbf{78.0}& \textbf{57.8}& \textbf{85.5} \\ 
    \bottomrule
    \end{tabular}
    \vspace{-5pt}
  \caption{
  {Ablation study on different category sets}.
  We report the Top1 and Top5 mean accuracy on classifying boxes and panoptic masks (thing and stuff)
  on COCO panoptic and ADE20K panoptic benchmarks.  %
  }
  \label{tab: catagory}
\end{table*}

\myPara{Ablation study on different category sets.}
To assess the impact of varying category sets, we conduct experiments using four various sets: 
133 (80), 171 (80), 273 (160), and 316 (160), categorized into foreground and background classes. 
The set of 133 categories exclusively comprises COCO Panoptic~\cite{lin2014microsoft} class set, 
while the 171-category set consists solely of COCO-Stuff~\cite{caesar2018coco} class set. 
The 273-category set integrates non-overlapping classes from both COCO-Stuff and the ADE20K Panoptic~\cite{zhou2017scene} dataset, which contains 150 categories. 
The 316-category set encompasses selected background classes from COCO-Stuff and the ADE20K dataset, which includes 847 categories. 
Our code accurately reflects all specified category sets.

As shown in~\tabref{tab: catagory},
with the increase in the number of categories, the performance of our model progressively improves 
when evaluated on the COCO Panoptic and ADE20K Panoptic benchmarks. 
This is because the larger category sets provide a richer representation of objects and stuff, 
enabling the model to capture more fine-grained information, 
thereby enhancing its overall performance.

\section{Visualizations}
\label{append: visualization}
\begin{figure*}[t]
    \centering
    \begin{overpic}[width=0.6\textwidth]{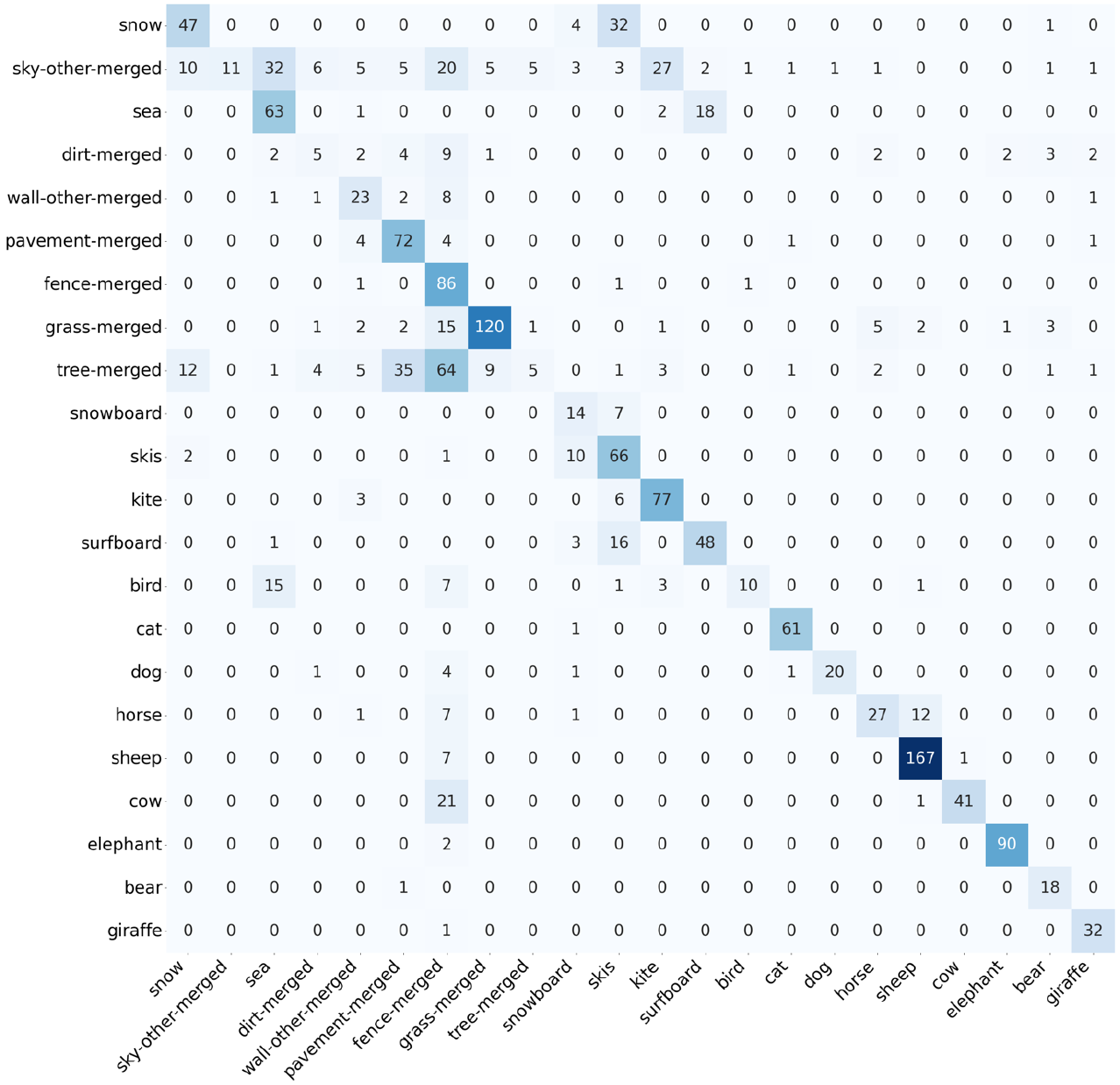}
        \put(0,60){\rotatebox{-90}{\small{Ground Truth}}}
        \put(50,1){\rotatebox{0}{\small{Prediction}}}

    \end{overpic}
    \caption{{Confusion matrix visualization for region classification results of EVA-CLIP.}
    }
    \label{fig: eav_conf_matrix}
\end{figure*}

\begin{figure*}[t]
    \centering
    \begin{overpic}[width=0.6\textwidth]{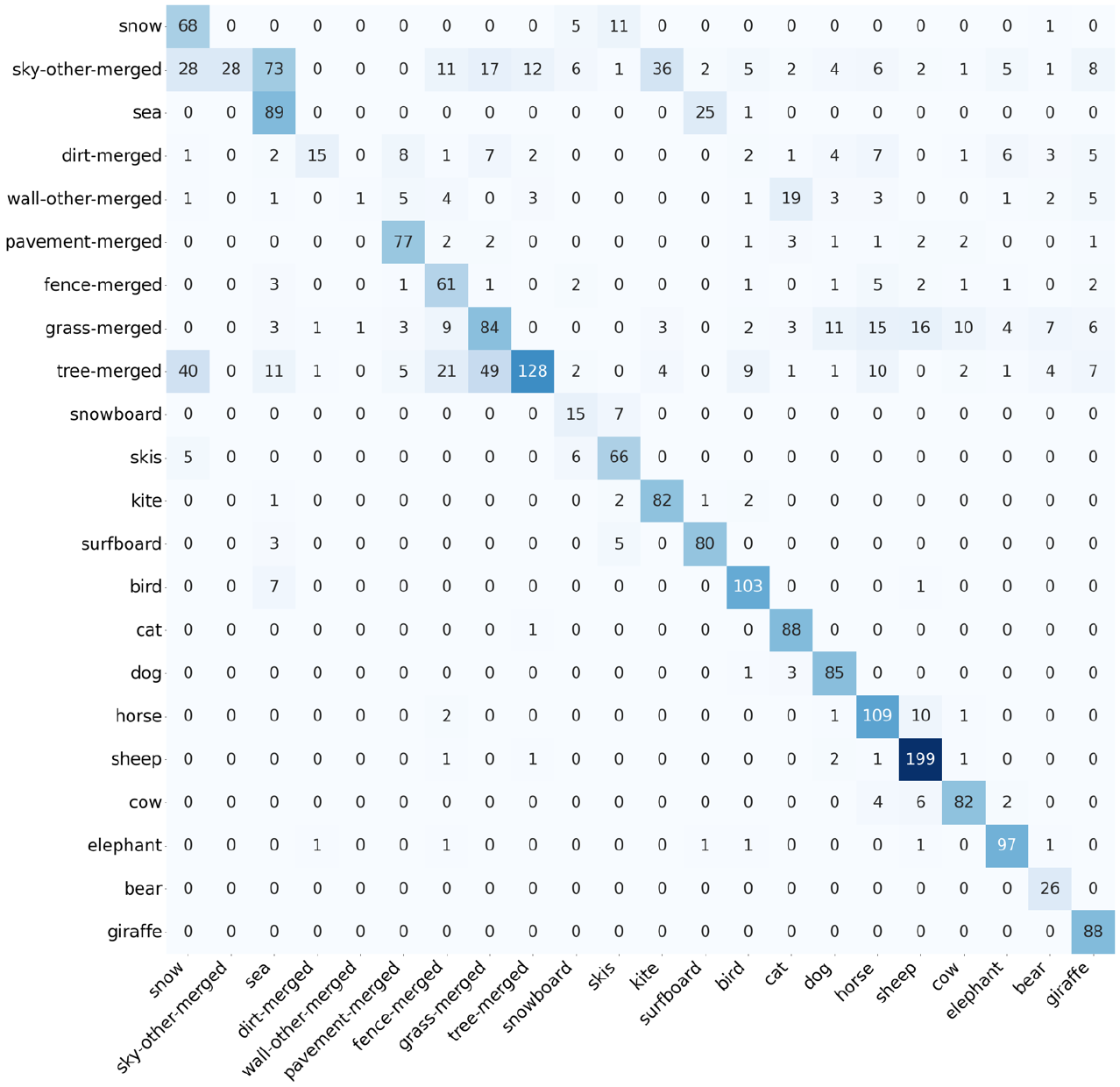}
        \put(0,60){\rotatebox{-90}{\small{Ground Truth}}}
        \put(50,1){\rotatebox{0}{\small{Prediction}}}
    \end{overpic}
    \caption{{Confusion matrix visualization for region classification results of RegionCLIP.}
    }
    \label{fig: regionclip_conf_matrix}
    \vspace{-7pt}
\end{figure*}

\begin{figure*}[t]
    \centering
    \begin{overpic}[width=0.6\textwidth]{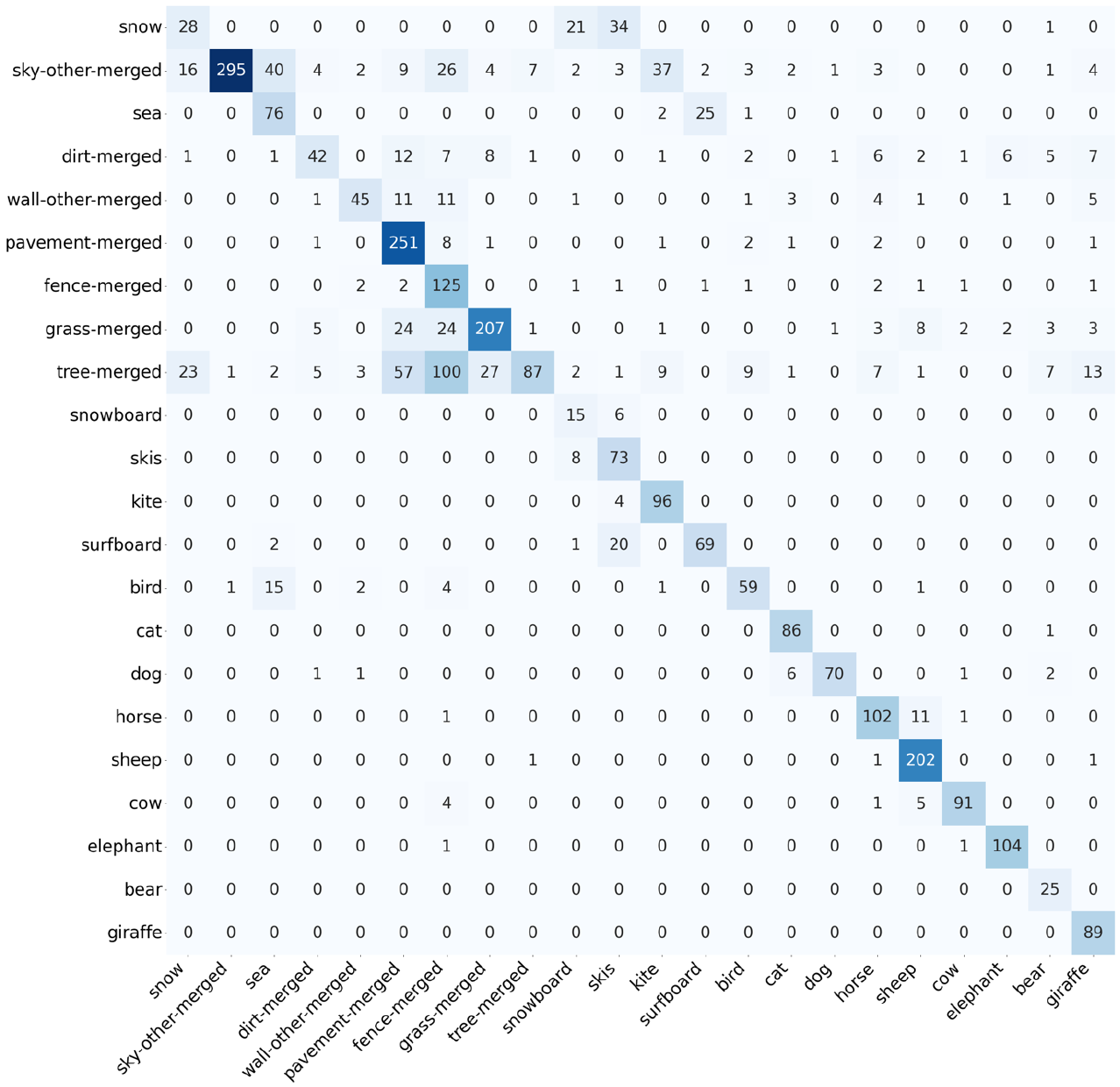}
        \put(0,60){\rotatebox{-90}{\small{Ground Truth}}}
        \put(50,1){\rotatebox{0}{\small{Prediction}}}
        \end{overpic}
    \caption{{Confusion matrix visualization for region classification results of CLIPSelf.}
    }
    \label{fig: clipself_conf_matrix}
\end{figure*}

\begin{figure*}[t]
    \centering
    \begin{overpic}[width=0.6\textwidth]{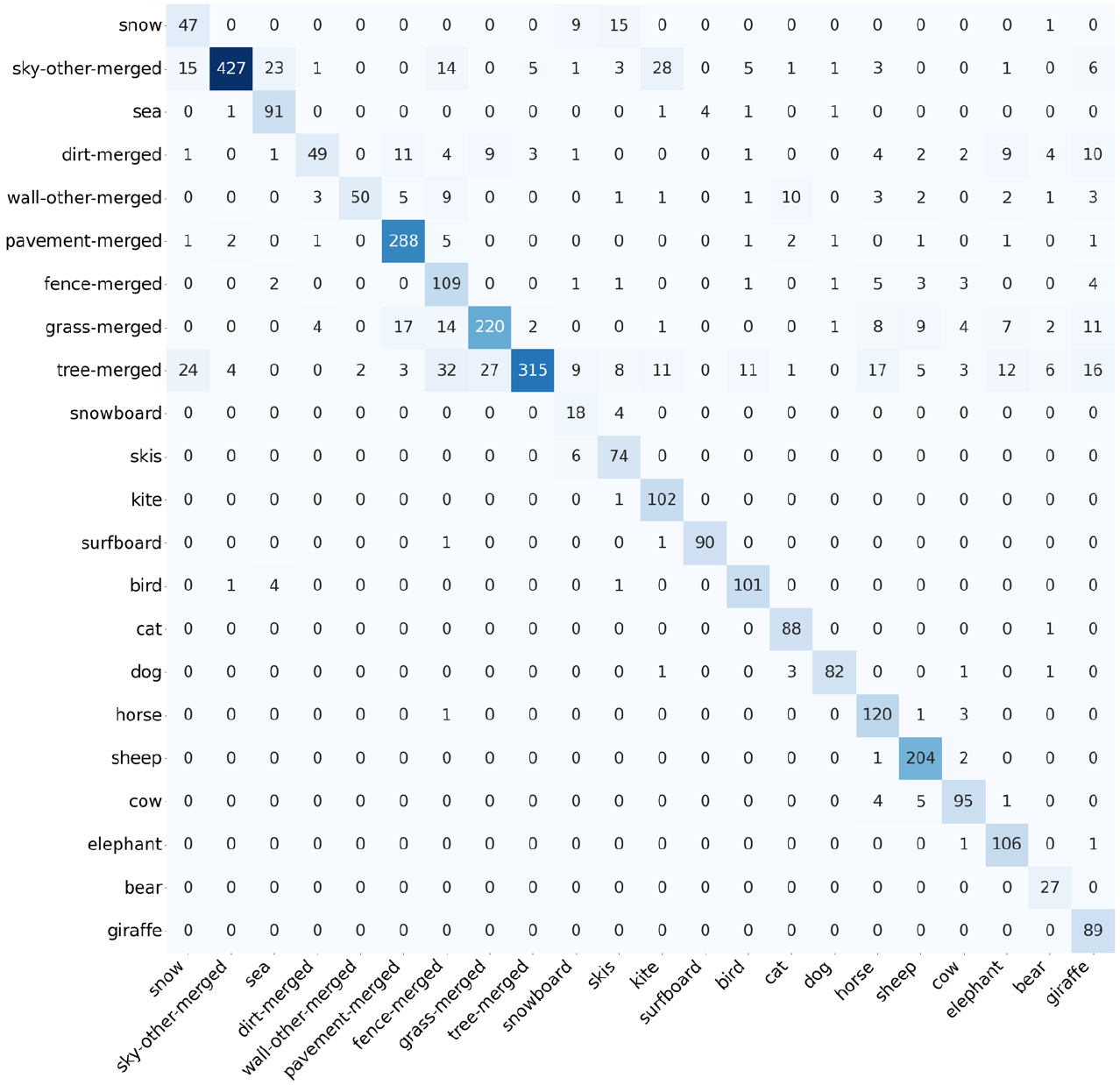}
        \put(0,60){\rotatebox{-90}{\small{Ground Truth}}}
        \put(50,1){\rotatebox{0}{\small{Prediction}}}
    \end{overpic}
    \caption{{Confusion matrix visualization for region classification results of \methodname.}
    }
    \label{fig: clipkd_conf_matrix}
    \vspace{-7pt}
\end{figure*}

\myPara{Confusion matrix.}
We compare the region classification results of our method against previous approaches
through a confusion matrix visualization 
for panoptic masks (both thing and stuff categories) on the COCO Panoptic dataset.
These confusion matrixes offer a systematic overview of region classification performance, 
illustrating the incidence of accurate and erroneous classifications, 
particularly facilitating a precise assessment of models' accuracy in differentiating between thing and stuff categories.
As shown in~\figref{fig: eav_conf_matrix},~\figref{fig: regionclip_conf_matrix}, and~\figref{fig: clipself_conf_matrix}, prior methods, 
including EVA-CLIP~\cite{sun2023eva}, RegionCLIP~\cite{zhong2022regionclip}, and CLIPSelf~\cite{wu2023clipself}, 
often misclassify background regions as co-occurring foreground classes, 
such as incorrectly identifying \textit{snow} as \textit{skis} or \textit{sky} as \textit{kite}.
In contrast, as demonstrated in~\figref{fig: clipkd_conf_matrix}, 
our \methodname{} achieves higher accuracy in recognizing each category, 
with a notable improvement in the precision of background object identification.

\begin{figure*}[!t]
    \centering
    \begin{overpic}[width=0.85\textwidth]{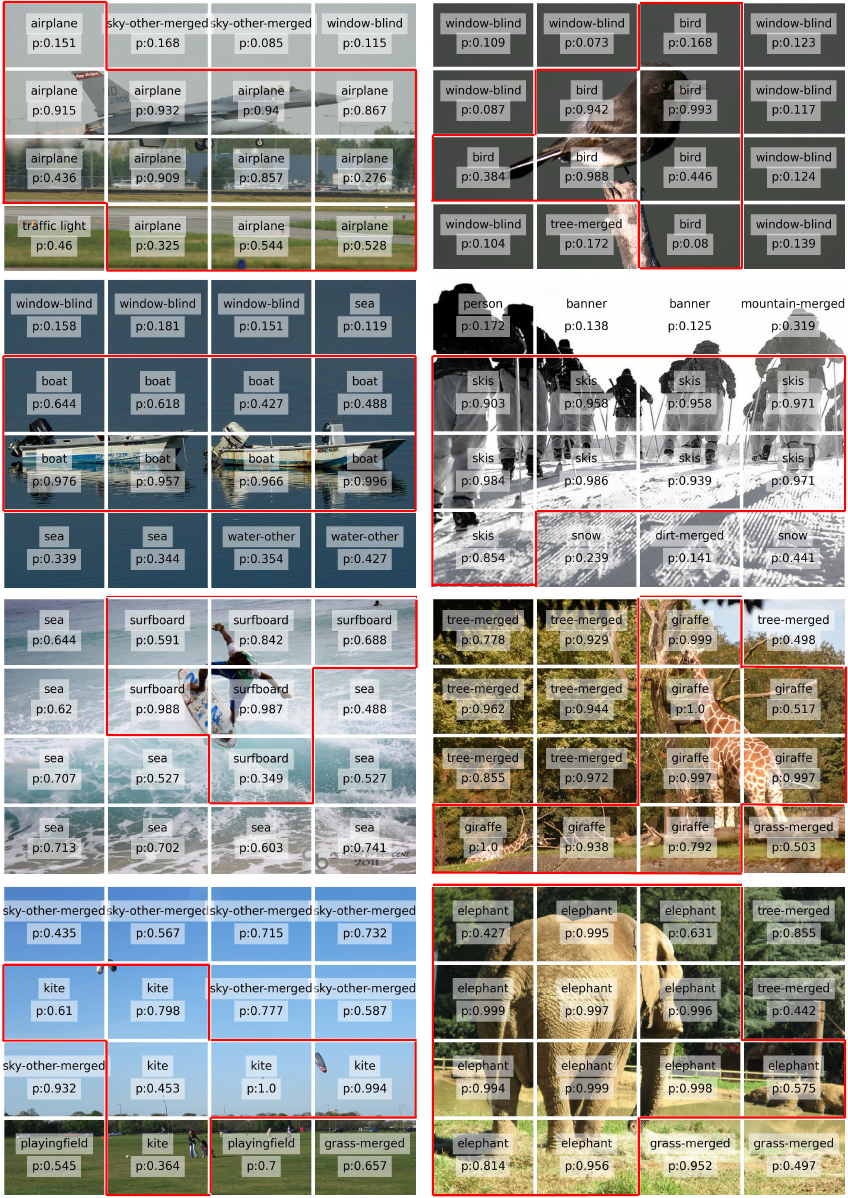}
    \end{overpic}
    \vspace{-5pt}
    \caption{{Visualization of image grid patches classification.}
    The powerful ViT-L/14 model exhibits a pronounced focus on foreground object recognition, 
    even when significant portions of background patches are misclassified as foreground objects.
    }
    \label{fig: forground_bias}
    \vspace{-7pt}
\end{figure*}

\myPara{Image grid patches classification.}
We visualize the classification results of image grid patches 
using the powerful ViT-L/14 model from CLIPSelf~\cite{wu2023clipself}.
As shown in~\figref{fig: forground_bias}, the model focuses heavily on foreground object recognition, 
but significant portions of background patches are misclassified as foreground objects.
The training VLMs are prone to learning these errors. 
Furthermore, regions with incorrect classifications often have low confidence scores, 
highlighting the importance of filtering them out.

\section{Datasets of training and evaluation}
\label{append: dataset}



\noindent\textbf{COCO:}
COCO~\cite{lin2014microsoft} is a large-scale panoptic segmentation dataset encompassing 80 \textit{Thing} and 53 \textit{Stuff} categories. 
The dataset comprises 118,000 images designated for the training set and 5,000 images for the validation set.

\noindent\textbf{ADE20k:}
ADE20k~\cite{ade20k} spans a broad spectrum of indoor and outdoor scenes, comprising 2,000 images for the validation set.
This dataset includes 100 \textit{Thing} and 50 \textit{Stuff} categories.
We evaluate open-vocabulary semantic annotations using both the extensive 847-category version (referred to as A-847) and the more frequently adopted 150-category version (referred to as A-150).

\noindent\textbf{Pascal Context:}
Pascal-Context~\cite{everingham2010pascal} constitutes an extensive dataset derived from Pascal-VOC 2010. 
We evaluate open-vocabulary semantic annotations using the complete set of 459 classes, 
referred to as PC-459.

\noindent\textbf{OV-COCO:}
The open-vocabulary detection COCO (OV-COCO) benchmark, introduced in OV-RCNN~\cite{Zareian_2021_CVPR}, 
divides the 65 object categories in the COCO dataset into 48 base categories and 17 novel categories.

\noindent\textbf{OV-LVIS:}
The open-vocabulary detection LVIS (OV-LVIS) benchmark, introduced in ViLD~\cite{gu2021open}, defines the 337 rare categories 
from LVIS v1.0 dataset~\cite{Gupta_2019_CVPR} as novel categories.

\end{document}